\pdfoutput=1
\documentclass[journal]{IEEEtran}
\usepackage{amsmath,amsfonts}
\usepackage{array}
\usepackage{booktabs}
\usepackage{longtable}
\usepackage{placeins}
\usepackage{graphicx}
\usepackage{xcolor}
\usepackage{url}
\usepackage{cite}
\graphicspath{{figures/}}

\begin{document}

\title{AgenticEvolve: Evolutionary Program Search with Persistent Language Model Agents that Design and Run Experiments}

\author{Ze-Yu Peng, Qing-Yu Lan, Zhan-He Wang, Jun Zhang, and Yun-Song Piao%
\thanks{This work has been submitted to the IEEE for possible publication. Copyright may be transferred without notice, after which this version may no longer be accessible.}%
\thanks{The work of J. Zhang was supported by the National Natural Science Foundation of China (NSFC) under Grant E414660101 and Grant 12547104, and by the Fundamental Research Funds for the Central Universities under Grant E4EQ6604X2 and Grant E3ER6601A2. The work of Y.-S. Piao was supported by the National Key Research and Development Program of China under Grant 2021YFC2203004 and by the NSFC under Grant 12475064.}%
\thanks{Z.-Y. Peng, Q.-Y. Lan, Z.-H. Wang, and Y.-S. Piao are with the School of Physical Sciences, University of Chinese Academy of Sciences, Beijing 100049, China.}%
\thanks{Z.-Y. Peng, J. Zhang, and Y.-S. Piao are with the International Centre for Theoretical Physics Asia-Pacific, University of Chinese Academy of Sciences, Beijing 100190, China.}%
\thanks{Z.-Y. Peng and J. Zhang are with the Taiji Laboratory for Gravitational Wave Universe, University of Chinese Academy of Sciences, Beijing 100049, China.}%
\thanks{Y.-S. Piao is with the School of Fundamental Physics and Mathematical Sciences, Hangzhou Institute for Advanced Study, University of Chinese Academy of Sciences, Hangzhou 310024, China, and with the Institute of Theoretical Physics, Chinese Academy of Sciences, Beijing 100190, China.}%
\thanks{Corresponding author: Ze-Yu Peng (e-mail: pengzeyu23@mails.ucas.ac.cn).}%
}

\markboth{}%
{Z.-Y. Peng \MakeLowercase{\textit{et al.}}: Evolutionary Program Search with Persistent Agents}

\maketitle

\begin{abstract}
    Evolutionary program search with large language models keeps the structure of an evolutionary algorithm and uses the model as its variation operator. Typically, the model generates each candidate program in a single request from parent programs and their scores. We present AgenticEvolve, an evolutionary program search framework in which each island is run by a persistent large language model agent. Between evaluations, the agent designs and runs experiments on the candidate program. To measure the contribution of experimentation, we compare AgenticEvolve with No-analysis, an otherwise identical configuration with experiments disabled. On two heuristic design tasks with three large language models, AgenticEvolve produces better programs than No-analysis and OpenEvolve on both hidden and unseen instances. It reaches the baselines' final quality with far fewer evaluations and comparable cost. On twelve equation discovery problems, it recovers the ground-truth structure of more equations than No-analysis and existing language-model and genetic programming baselines.
\end{abstract}


\section{Introduction}
Genetic programming searches over programs with variation operators that recombine and mutate code without knowledge of what it computes~\cite{koza1992}. Evolutionary program search with large language models (LLMs) retains the structure of the evolutionary algorithm and uses the model as its variation operator~\cite{wu2025llm}. The search maintains a population of programs, prompts the model with one or more parents and their scores, and evaluates each returned program with an automated evaluator. FunSearch uses this approach to find new constructions in extremal combinatorics and a new online bin packing heuristic~\cite{funsearch}, and AlphaEvolve extends it to algorithm and systems optimization~\cite{alphaevolve}. Evolution of Heuristics and LLaMEA apply it to heuristic and metaheuristic design~\cite{eoh,llamea}. LLM-SR uses the same loop to discover scientific equations, represented as programs whose constants are fitted numerically~\cite{llmsr}.

In these LLM-based systems, however, the model has no execution tools available during candidate generation. It reasons from the parent programs, their scores, and any textual feedback supplied in the prompt, such as reflections on earlier evaluations. Empirical information about program behavior enters the search only through formal evaluation. Each formal evaluation consumes budget, and its instances are often hidden from the model.

Recent advances in LLM agents have enabled models to carry out multi-step tasks by interleaving reasoning with tool calls~\cite{react}. Coding agents use tools to edit files, run code, and inspect outputs iteratively to optimize program performance~\cite{sweagent,algotune}. Within evolutionary search, such an agent can experiment between formal evaluations: it can generate instances, measure program behavior, fit parameters and test changes locally before submitting a candidate. These experiments can support local improvement and surrogate evaluation, both of which have long been combined with global search in evolutionary computation~\cite{memetic,surrogate}. Recent work has also used persistent agents for program search~\cite{coral} and ablated diagnostic or data-analysis tools in heuristic design and equation discovery~\cite{ahdagent,srscientist}. In this paper, we ask how much self-designed experiments improve program quality when the other evolutionary mechanisms remain fixed, what resources they require, and how their benefits and costs vary across tasks and models.

This paper presents AgenticEvolve, an evolutionary program search framework in which each island is run by a persistent LLM agent. Throughout the search, the agent maintains a conversation in which it edits the candidate program, writes and executes custom analysis scripts between commits, and submits candidates for formal evaluation through a commit tool. Committed programs enter a MAP-Elites archive~\cite{mapelites}, and islands exchange elite programs through periodic migration. A No-analysis configuration removes the analysis tool and leaves everything else unchanged, which isolates the contribution of the agent's experiments. We evaluate this contribution across two regimes: heuristic design on online bin packing and dynamic job shop scheduling, where formal evaluation instances are hidden and the agent must generate its own test instances, and scientific equation discovery on twelve problems from LLM-SRBench~\cite{llmsrbench}, where training data are available for direct analysis.

The paper makes three contributions:
\begin{itemize}
    \item An evolutionary program search framework in which persistent LLM agents conduct island-based search and run self-designed experiments between formal evaluations, paired with a matched No-analysis configuration that isolates the effect of experimentation.
    \item An empirical evaluation on two heuristic design tasks across three LLMs against No-analysis and OpenEvolve~\cite{openevolve}. AgenticEvolve achieves the best mean final quality on both hidden evaluation instances and unseen test instances with every model, reaching the baselines' final quality with a fraction of their formal evaluations, while its cost advantage depends on the model and budget.
    \item A cross-task evaluation on scientific equation discovery against No-analysis, LLM-SR~\cite{llmsr}, SR-Scientist~\cite{srscientist}, and PySR~\cite{pysr}. Under common criteria for structure recovery, AgenticEvolve recovers more ground-truth equations than all baselines at a cost comparable to that of SR-Scientist. A case study illustrates how experiments help construct expressions and determine difficult constants.
\end{itemize}
The source code is available at \url{https://github.com/pengzy-cosmo/AgenticEvolve}.

\section{Related Work}
\subsection{Evolutionary Program Search}
FunSearch, AlphaEvolve and OpenEvolve search over programs with an evolutionary outer loop and an LLM as the variation operator. FunSearch keeps a database of programs organized in islands and samples high-scoring programs into the prompt~\cite{funsearch}. AlphaEvolve extends the search to open mathematical problems and to the optimization of compute stacks~\cite{alphaevolve}. OpenEvolve, the baseline in Section~\ref{sec:heuristics}, is an open-source system that follows this approach, with a MAP-Elites archive, multiple islands with periodic migration, and prompts containing parent and inspiration programs~\cite{openevolve}. For heuristic design, Evolution of Heuristics evolves natural-language ideas together with code under fixed prompt strategies~\cite{eoh}, ReEvo adds reflections on evaluation results~\cite{reevo}, and LLaMEA mutates complete metaheuristics in a (1+1) scheme driven by performance feedback~\cite{llamea}. ShinkaEvolve refines the outer loop through parent sampling, novelty-based candidate rejection and adaptive model selection~\cite{shinkaevolve}. In their default configurations, these systems generate each candidate in a single model request without execution tools. AgenticEvolve retains population-based search, archives and islands, while a persistent agent develops candidates through experiments between formal evaluations.

\subsection{Agentic Search}
Agentic search methods differ in how much control they give the agent over the search. AlgoTune uses a general coding agent to speed up numerical programs without maintaining a population~\cite{algotune}. CORAL runs persistent coding agents against a hidden grader. The agents can inspect all previous attempts and choose their own parents. The study reports a correlation between local verification and improvement, but does not ablate execution~\cite{coral}. AVO uses an autonomous coding agent as the variation operator in evolutionary search and evaluates it in a multi-day run on attention-kernel optimization~\cite{avo}. ReASearch and SimpleEvol remove the external search controller, using a single agent that decides what to evaluate~\cite{reasearch} and an agent loop with minimal human priors~\cite{simpleevol}, respectively. AgenticEvolve combines persistent agents with population-based search, MAP-Elites archives and periodic migration.

These methods also differ in the information agents can obtain between evaluations. AHD Agent provides two predefined read-only diagnostic tools, and an ablation in its appendix shows that they lower the agent's optimality gap by one to two points~\cite{ahdagent}. Removing ReASearch's Python execution tool reduces its accuracy on ARC-AGI-2 by a third~\cite{reasearch}. Vesper places coding agents with file-system access inside evolutionary search. At a fixed token budget, it finds that allocating more reasoning to fewer candidates outperforms generating more candidates~\cite{vesper}. In AgenticEvolve, the model writes its own experiments for the current candidate. We measure their contribution through the No-analysis configuration, which disables this capability while retaining the other search settings, and report both model costs and formal evaluation counts.

\subsection{Scientific Equation Discovery}
Scientific equation discovery is commonly cast as symbolic regression, and research on it spans genetic programming, LLM-based program evolution and agentic search. PySR searches expression trees with multi-population genetic programming~\cite{pysr}, and LaSR adds LLM-guided operators and a library of natural-language concepts to this search~\cite{lasr}. LLM-SR uses a language model to propose equation skeletons as programs and fits their constants numerically~\cite{llmsr}. Deliberate Evolution adds structure diagnosis, adaptive operators and reflective memory to this loop~\cite{deliberateevolution}. SR-Scientist equips a single agent with a data analysis tool and an equation evaluator, and maintains a memory of candidate equations~\cite{srscientist}. Removing the data analysis tool reduces the fraction of problems fitted within a tolerance of 0.01 from 63.6\% to 35.7\% for one model, with smaller reductions for two others. LLM-SRBench, from which our problems are drawn, evaluates numerical accuracy and uses an LLM to judge symbolic accuracy~\cite{llmsrbench}. Section~\ref{sec:sr} compares AgenticEvolve and No-analysis with PySR, LLM-SR and SR-Scientist.

\section{AgenticEvolve}
\begin{figure*}[t]
    \centering
    \includegraphics[width=\textwidth]{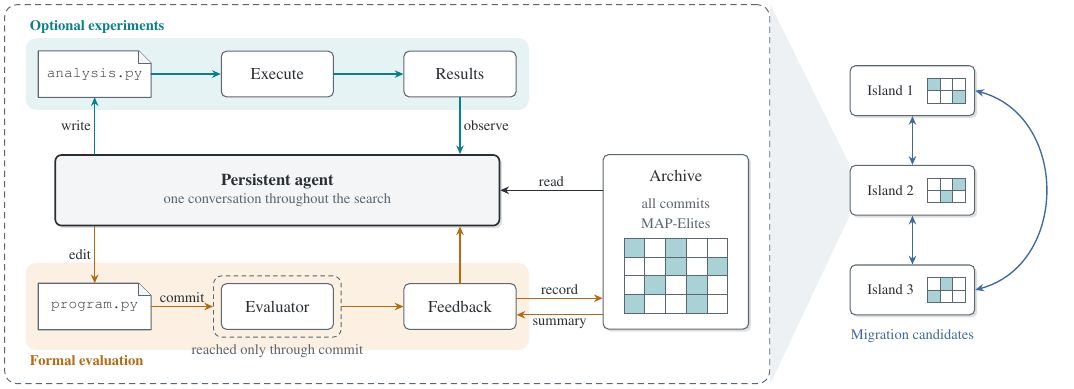}
    \caption{AgenticEvolve. Left: one island, magnified. A persistent agent edits \texttt{program.py} and commits it for formal evaluation. Between commits, it may write and execute \texttt{analysis.py} to run experiments. No-analysis removes these experiments. The feedback on each commit includes a summary of the archive, and the agent can read any archived program, including those of other islands. Right: islands run concurrently and exchange migration candidates, which appear in the archive summary. The agent decides whether to use them.}
    \label{fig:architecture}
\end{figure*}
AgenticEvolve organizes program search across multiple concurrently running islands (Fig.~\ref{fig:architecture}). Each island is run by a persistent LLM agent, which maintains a conversation with the framework throughout the search. The agent uses tools to inspect and modify a candidate program, to run experiments on it and to commit it for formal evaluation. The framework evaluates every committed program, records it in an archive and returns the result to the agent together with a summary of the archive, which periodically includes elite programs from the neighboring islands.

\subsection{Problem and Task Interface}\label{sec:task}
AgenticEvolve searches for a program that maximizes fitness within a fixed budget. A task is specified by a natural-language description of the problem, an initial program and an evaluator. The evaluator executes a candidate and returns its fitness together with task-defined features, which determine its position in the archive of Section~\ref{sec:archive}. A task may also provide data for the agent's experiments. Every island starts from the same initial program. At the end of a run, the framework returns the program with the highest fitness among the successfully evaluated commits from all islands.

\subsection{One Island: Persistent Conversation and Experiments}\label{sec:island}
The conversation of an island begins with a system prompt that describes the task, explains the feedback the agent will receive and lists the files and tools available to it. The first message contains the initial program, its evaluation result and the budget. Afterwards the framework only returns the results of tool calls and prompts the agent to continue whenever it ends a turn. The accumulated conversation thus provides the agent with a record of previous program revisions, experiments and evaluation results on which to base subsequent decisions.

The agent inspects and modifies the candidate with three file tools, \texttt{read}, \texttt{write} and \texttt{edit}, which operate on \texttt{program.py}. Between two commits it can also run experiments on the candidate. It writes a second file, \texttt{analysis.py}, which can import the current program, and executes this script with the tool \texttt{run\_analysis}. We refer to each such execution as an analysis. The framework runs the script in a working directory that contains the program, the script and the data provided by the task, if any, and returns the printed output to the conversation. The evaluator and its hidden inputs are not available in this directory. After reading the output, the agent revises the program or the script, runs another analysis, or commits the program. The agent decides what to investigate, how many analyses to run and what to change, subject only to a limit on the number of analyses per commit.

The tool \texttt{commit} submits the current \texttt{program.py} for formal evaluation. Each submission consumes one unit of the evaluation budget, including submissions that fail. The feedback contains the evaluation result, whose content is restricted as described in Section~\ref{sec:protocol}, and a summary of the archive described in Section~\ref{sec:archive}. Programs appear in this summary as one-line entries, and the agent can retrieve the complete code of any archived program, including those of other islands, through the tool \texttt{read\_commit} and continue from it.

To keep the conversation within the context window of the LLM, the framework compacts it when its length reaches a configured fraction of that window. At the framework's request the agent summarizes the search history, including unsuccessful approaches, experimental findings and its current plan. The conversation then continues from this summary together with the current program and the latest feedback. The candidate program, the analysis script, the archive and the accumulated budget counters are retained through compaction.

\subsection{Archive and Islands}\label{sec:archive}
The archive records every commit from all islands, including failed evaluations, with its source code, evaluation result and the description written by the agent. Each island maintains a MAP-Elites grid~\cite{mapelites} over the features computed by the evaluator, such as program complexity. Each feature is divided into uniform bins, and a successfully evaluated program is assigned to a cell according to its feature values. The program enters the cell if the cell is empty or if its fitness is higher than that of the current occupant. The archive summary returned after each commit lists the island's best program and its grid elites, each given by its identifier, metrics and description.

The islands are arranged in a ring and exchange elite programs through periodic migration~\cite{islandmodel}. After a fixed number of commits by an island, the framework pools the grid elites of its two neighbors and selects a configured fraction with the highest fitness. Those not previously shown to the island are added to its next archive summary as migration candidates, in the same format as the local elites and with their island of origin identified. The agent decides whether to retrieve a candidate through \texttt{read\_commit} and use it to revise or replace its current program.

\subsection{Evaluation Protocol and the No-analysis Configuration}\label{sec:protocol}
The evaluation result returned by \texttt{commit} is restricted. A successful evaluation returns the task-defined metrics. A failed evaluation returns the exception type and its location in the program, with a diagnostic message when the evaluator itself rejects the program. Raw evaluation output, including printed text and full tracebacks, is stored in the archive but withheld from the agent. This restriction prevents candidates from using printed output or exception messages to return arbitrary results computed during evaluation, such as statistics of the evaluation data. Execution follows protocol-level access restrictions and uses no operating-system sandbox, a limitation discussed in Section~\ref{sec:discussion}.

We compare AgenticEvolve with No-analysis, an otherwise identical configuration without experiments. No-analysis removes \texttt{run\_analysis} from the available tools, disables access to \texttt{analysis.py} and removes the corresponding instructions from the system prompt. The persistent conversation, program editing, formal evaluation, archive, migration, compaction and budget remain unchanged. Candidate programs are still executed at each commit, with the restricted feedback described above. The difference between the two configurations therefore measures the contribution of the agent's experiments.

\section{Experimental Design}
\label{sec:design}
We evaluate AgenticEvolve on two heuristic design tasks, online bin packing (OBP) and dynamic job shop scheduling (DJSS). We chose these tasks because the instances used for formal evaluation are hidden from the agent, while the task description specifies their generating process. The agent's experiments therefore cannot reproduce the formal evaluation, and any benefit of analysis must come from instances that the agent generates itself. In Section~\ref{sec:sr} we extend the study to symbolic regression, where the agent is given the training data and its experiments can reproduce the training score.

\subsection{Tasks}
OBP follows the setting of FunSearch~\cite{funsearch}. Items arrive one at a time and must be placed irrevocably. The candidate program scores the feasible bins for the current item, and the item is placed in the bin with the highest score. The initial program is Best Fit, which places the item in the feasible bin with the least remaining capacity. The description states the instance distribution, sequences of 120 items with integer sizes drawn uniformly from $[20,100]$ and bins of capacity 150. The hidden evaluation set contains 100 such sequences. The evaluator computes the fractional excess $e$ of the total number of bins used over the sum of the Martello--Toth $L_2$ lower bounds~\cite{martello1990} and returns the fitness $1/(1+e)$. We report quality as this excess in percent. Generalization is tested on the 20 instances of OR1 from the OR-Library~\cite{beasley1990}, which follow the same size distribution, and on 100 new sequences drawn from the stated distribution. Best Fit and the heuristic reported by FunSearch serve as references.

DJSS follows the simulation setting of Ferreira et al.~\cite{ferreira2022}. Jobs arrive according to a Poisson process at a shop with ten machines. Each job has two to fourteen operations with integer processing times drawn uniformly from $[1,99]$, and consecutive operations visit different machines. Its due date is the arrival time plus an allowance factor times the total processing time. The candidate program is a dispatching rule. Whenever a machine becomes free, the rule scores its waiting operations based on job attributes and the current shop state. The operation with the highest score starts next. The initial program is the shortest processing time rule. The evaluation uses five scenarios combining utilization 0.85 with allowance factors 2 and 4, and utilization 0.95 with factors 2, 4 and 6. Each scenario uses ten hidden seeds. The combination of utilization 0.85 and allowance 6 is excluded because its tardiness is close to zero. In each simulation, the first 500 jobs form the warm-up period, and the mean tardiness of the following 2000 jobs is recorded. For each scenario, this value is averaged over the seeds and divided by the corresponding mean for rule R1 of Ferreira et al. Quality is the geometric mean of these ratios over the five scenarios, and the fitness is its reciprocal. Generalization is tested on the same five scenarios with ten new seeds, so it covers new arrival streams but not new scenarios. R1 is the reference and has quality 1 by construction.

\subsection{Methods and Models}
AgenticEvolve may run up to ten analyses per commit. No-analysis is the configuration of Section~\ref{sec:protocol}. The baseline is OpenEvolve~\cite{openevolve}, an open-source implementation of AlphaEvolve~\cite{alphaevolve}. In each iteration OpenEvolve samples a parent and several reference programs from the current island and asks the LLM to modify the parent in a single request. We retain this search mechanism and align the remaining settings with AgenticEvolve. Both systems use three islands, the same MAP-Elites features with ten bins each, the same task description and initial program, the same evaluator and executor with the same evaluation time limit, and the same model, reasoning effort and output limit. OpenEvolve receives the task description as its system message. Changes to OpenEvolve's default settings, such as a migration interval short enough to take effect within a run, are listed and justified in Section S1-B of the supplementary material.

Each method is run with DeepSeek-V4.1-Flash, GLM-5.3-Flash and Qwen3.8-Flash. Table~\ref{tab:models} gives the prices and reasoning effort of these three LLMs.

\begin{table}[t]
    \centering
    \caption{LLMs Used in the Experiments, with List Prices in USD per Million Tokens and Reasoning Effort}
    \label{tab:models}
    \begin{tabular}{lrrrl}
        \toprule
        Model & Input & Cached input & Output & Reasoning \\
        \midrule
        DeepSeek-V4.1-Flash & 0.15 & 0.003 & 0.60 & high \\
        GLM-5.3-Flash & 0.15 & 0.03 & 0.50 & high \\
        Qwen3.8-Flash & 0.15 & 0.016 & 0.47 & medium \\
        \bottomrule
    \end{tabular}
\end{table}

\subsection{Budgets, Resource Measures and Statistics}
Each island of AgenticEvolve and No-analysis commits 30 times, so a run evaluates 90 candidates in addition to the initial program. OpenEvolve runs 180 iterations and evaluates up to 180 candidates, so that its curve extends beyond this budget. Because the budgets differ, OpenEvolve is compared with the other two methods at common budgets on each resource axis, and the final values of the complete runs are reported in addition.

Results are reported against two resource measures. The number of formal evaluations counts every call of the evaluator, including the initial program and evaluations that fail or time out. The estimated LLM cost is computed from the recorded token usage of the requests at the prices of Table~\ref{tab:models}.

Each method is run five times with each model. The runs are independent, so all comparisons are unpaired. Differences in final quality are tested with a two-sided Mann--Whitney $U$ test at the 0.05 level, using an exact calculation when there are no ties and an asymptotic approximation otherwise. With five runs per group, the smallest attainable exact $p$ value is 0.008. Effect sizes are reported as differences in means with bootstrap 95\% confidence intervals. We report $p$ values without correction. A Holm correction applied to the three comparisons within each task and model would not alter the significance of any difference in final quality.

Tasks, budgets and stopping conditions were fixed before the formal runs, and a run was repeated only after an infrastructure failure. The heuristic runs predate the feedback restriction of Section~\ref{sec:protocol}, so for all three methods the feedback of a failed evaluation included the truncated output and traceback. An audit of these runs (Section S4-A) found no access to the evaluator or the hidden instances.

\section{Heuristic Optimization}\label{sec:heuristics}

\subsection{Final Quality and Generalization}
\begin{figure*}[t]
    \centering
    \includegraphics[width=\textwidth]{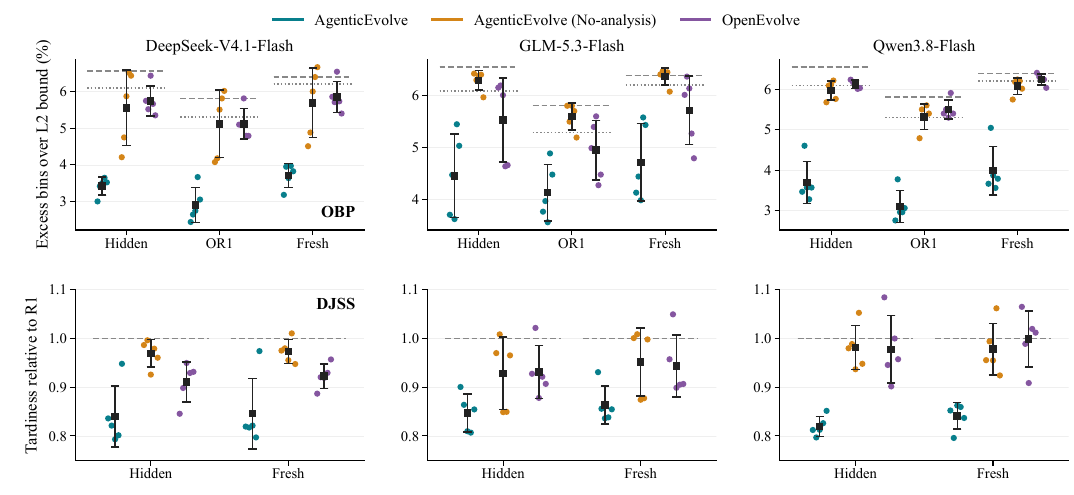}
    \caption{Final program quality on the hidden evaluation set and on unseen instances. The unseen OBP instances are OR1 and fresh sequences from the training distribution, and the unseen DJSS instances are the same scenarios with fresh seeds. Each point is one run. Black squares and bars are the mean $\pm$ SD over five runs. Gray dashes mark Best Fit, the FunSearch heuristic and rule R1.}
    \label{fig:quality}
\end{figure*}

Table~\ref{tab:heuristics} and Fig.~\ref{fig:quality} report the quality of each run's final program on the hidden evaluation set and on unseen instances. On the hidden set, AgenticEvolve has the best mean quality with every model on both tasks. Across the three models, it reduces the OBP excess over the $L_2$ bound by 1.8 to 2.3 percentage points compared with No-analysis and by 1.1 to 2.4 points compared with OpenEvolve. On DJSS it reduces R1-normalized tardiness by 0.08 to 0.16 compared with No-analysis and by 0.07 to 0.16 compared with OpenEvolve. The confidence intervals of all twelve differences exclude zero (Section S2-A). Nine of the differences are significant, the exceptions being No-analysis on DJSS with GLM and OpenEvolve on DJSS with DeepSeek and on OBP with GLM. No difference between No-analysis and OpenEvolve is significant with any model on either task.

The improvements carry over to instances that no method has seen. AgenticEvolve has the lowest mean on every unseen set with every model, and the confidence interval of its difference from either other method excludes zero in every case (Section S2-A). On OR1, AgenticEvolve's mean excess over the $L_2$ bound ranges from 2.9 to 4.1 percent across models, compared with 5.3 percent for the FunSearch heuristic and 5.8 percent for Best Fit. The corresponding means for No-analysis and OpenEvolve range from 5.0 to 5.6 percent, close to the FunSearch heuristic. The fresh sequences from the training distribution give the same picture. On the fresh DJSS seeds, AgenticEvolve achieves mean R1-normalized tardiness of 0.84 to 0.86 across models, compared with 0.92 to 1.00 for the other two methods.

\begin{table*}[t]
\centering
\caption{Heuristic design results: mean $\pm$ SD over five runs of the final program's quality on the hidden set and on unseen instances (lower is better), and estimated cost per run. $p$: two-sided Mann--Whitney test of hidden-set quality against AgenticEvolve.}
\label{tab:heuristics}
\footnotesize
\setlength{\tabcolsep}{8pt}
\begin{tabular}{llrrrr}
\toprule
Model & Method & Hidden & Unseen & Cost (\$) & $p$ \\
\midrule
\multicolumn{6}{l}{\emph{Online bin packing (unseen: OR1)}} \\
\addlinespace[1pt]
DeepSeek & AgenticEvolve & 3.42 $\pm$ 0.25 & 2.92 $\pm$ 0.48 & 1.08 & -- \\
 & AgenticEvolve (No-analysis) & 5.54 $\pm$ 1.05 & 5.12 $\pm$ 0.92 & 0.42 & $<$0.01 \\
 & OpenEvolve & 5.74 $\pm$ 0.41 & 5.12 $\pm$ 0.42 & 3.06 & $<$0.01 \\
\addlinespace[2pt]
GLM & AgenticEvolve & 4.46 $\pm$ 0.80 & 4.14 $\pm$ 0.54 & 4.48 & -- \\
 & AgenticEvolve (No-analysis) & 6.30 $\pm$ 0.19 & 5.61 $\pm$ 0.26 & 0.56 & $<$0.01 \\
 & OpenEvolve & 5.54 $\pm$ 0.81 & 4.95 $\pm$ 0.57 & 0.69 & 0.10 \\
\addlinespace[2pt]
Qwen & AgenticEvolve & 3.70 $\pm$ 0.52 & 3.10 $\pm$ 0.39 & 3.69 & -- \\
 & AgenticEvolve (No-analysis) & 5.98 $\pm$ 0.24 & 5.32 $\pm$ 0.32 & 0.34 & 0.01 \\
 & OpenEvolve & 6.13 $\pm$ 0.10 & 5.50 $\pm$ 0.24 & 1.10 & 0.01 \\
\midrule
\multicolumn{6}{l}{\emph{Dynamic job shop (unseen: fresh seeds)}} \\
\addlinespace[1pt]
DeepSeek & AgenticEvolve & 0.840 $\pm$ 0.063 & 0.846 $\pm$ 0.072 & 0.60 & -- \\
 & AgenticEvolve (No-analysis) & 0.970 $\pm$ 0.028 & 0.974 $\pm$ 0.025 & 0.10 & 0.02 \\
 & OpenEvolve & 0.911 $\pm$ 0.041 & 0.923 $\pm$ 0.025 & 1.84 & 0.10 \\
\addlinespace[2pt]
GLM & AgenticEvolve & 0.847 $\pm$ 0.039 & 0.863 $\pm$ 0.039 & 1.47 & -- \\
 & AgenticEvolve (No-analysis) & 0.928 $\pm$ 0.074 & 0.952 $\pm$ 0.069 & 0.14 & 0.22 \\
 & OpenEvolve & 0.931 $\pm$ 0.054 & 0.944 $\pm$ 0.064 & 0.48 & 0.02 \\
\addlinespace[2pt]
Qwen & AgenticEvolve & 0.820 $\pm$ 0.020 & 0.841 $\pm$ 0.027 & 1.72 & -- \\
 & AgenticEvolve (No-analysis) & 0.981 $\pm$ 0.045 & 0.978 $\pm$ 0.053 & 0.13 & $<$0.01 \\
 & OpenEvolve & 0.978 $\pm$ 0.069 & 0.999 $\pm$ 0.057 & 0.64 & $<$0.01 \\
\bottomrule
\end{tabular}
\end{table*}

\subsection{Quality Against Formal Evaluations}
\begin{figure*}[t]
    \centering
    \includegraphics[width=\textwidth]{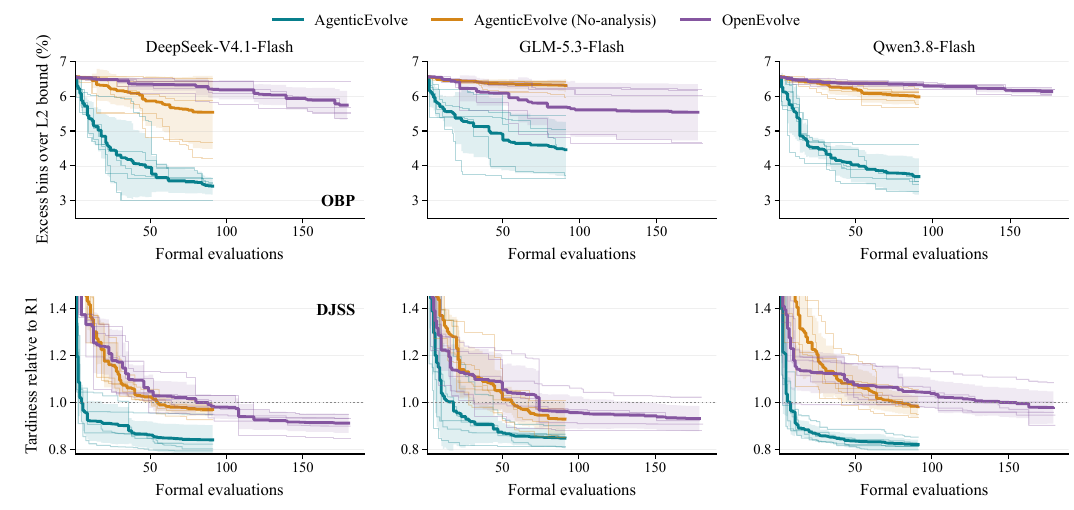}
    \caption{Best-so-far quality against formal evaluations. Thin lines are individual runs. Bold lines and bands are the mean $\pm$ SD over the common evaluation range. The initial job-shop program scores 1.648 and lies above the plotted range.}
    \label{fig:evaluations}
\end{figure*}

Fig.~\ref{fig:evaluations} plots the best quality found so far against the number of formal evaluations. The horizontal axis does not count analyses. AgenticEvolve uses formal evaluations most efficiently on both tasks with every model. Its curve separates from the other two within the first ten evaluations and stays ahead to the end of its budget. In five of the six cases, AgenticEvolve reaches OpenEvolve's final quality within 20 evaluations, compared with OpenEvolve's 180-iteration runs (Section S2-B). No-analysis and OpenEvolve have no separate analysis tool, and their curves remain close over the common range of 91 evaluations. On DJSS the two curves overlap with all three models. On OBP, No-analysis is slightly ahead of OpenEvolve with DeepSeek and Qwen, and slightly behind with GLM.

\subsection{Quality Against Cost}
\begin{figure*}[t]
    \centering
    \includegraphics[width=\textwidth]{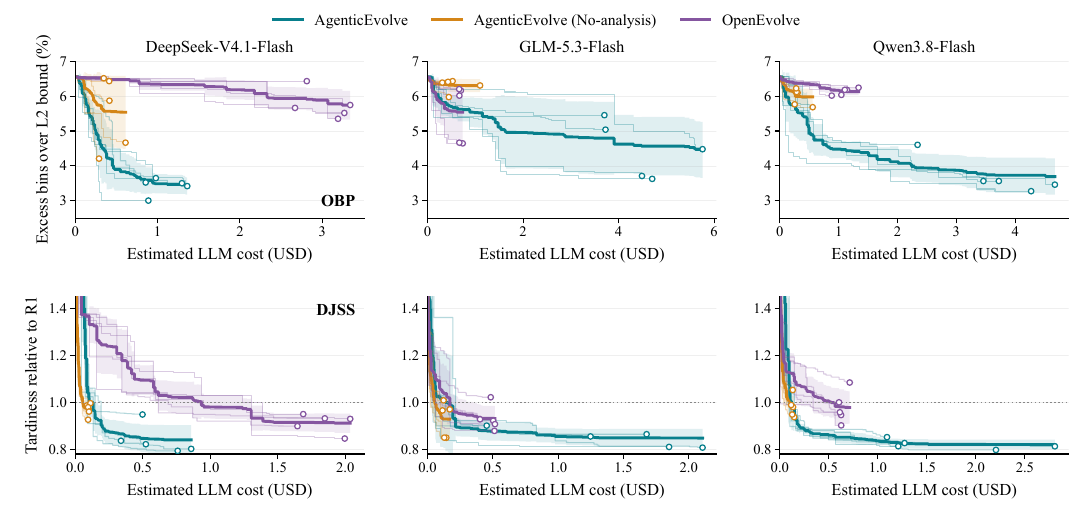}
    \caption{Best-so-far quality against estimated cost. Thin lines are individual runs. Bold lines and bands are the mean $\pm$ SD over the common evaluation range. Circles mark each run's actual spend. The initial job-shop program lies above the plotted range.}
    \label{fig:cost}
\end{figure*}

Fig.~\ref{fig:cost} plots best-so-far quality against estimated LLM cost. Although the analysis makes AgenticEvolve several times as expensive as No-analysis, its mean LLM cost is at most 4.5 USD per run across all model--task combinations (Table~\ref{tab:heuristics}). Relative to OpenEvolve, AgenticEvolve costs less with DeepSeek and more with GLM and Qwen.

At equal cost, AgenticEvolve achieves better quality than OpenEvolve with DeepSeek and Qwen. It reaches OpenEvolve's final quality for 0.08 to 0.19 USD, compared with 0.64 to 3.06 USD for a complete OpenEvolve run (Section S2-B), and maintains its advantage at larger budgets. With GLM, the two methods achieve similar quality over the cost range covered by OpenEvolve: their curves nearly coincide on OBP, while AgenticEvolve has a slight advantage on DJSS.

The equal-cost comparison with No-analysis depends on the task. On OBP, AgenticEvolve achieves better quality with every model, and the gap widens as the budget grows. On DJSS, No-analysis makes inexpensive improvements in its first few commits, while AgenticEvolve spends additional requests on analysis before submitting candidates. It catches up near the cost of a completed No-analysis run with Qwen and at a higher cost with DeepSeek and GLM.

The composition of the cost explains the dependence on the model. With DeepSeek, OpenEvolve generates an average of 15 to 27 thousand output tokens per request, mostly from extended reasoning in the absence of execution tools, compared with about one thousand for AgenticEvolve. This output dominates OpenEvolve's cost. Its responses are much shorter with Qwen (4.6 to 11.1 thousand tokens per request) and GLM (2.5 to 4.8 thousand), and OpenEvolve is cheaper with these models. AgenticEvolve incurs input costs throughout the search as it repeatedly sends the growing conversation. Although 99\% of these input tokens are cached, the higher cache-read prices of Qwen and GLM make repeated input the dominant expense, accounting for approximately 78\% and 87\% of AgenticEvolve's cost, respectively.

\begin{figure}[htbp]
    \centering
    \includegraphics[width=\columnwidth]{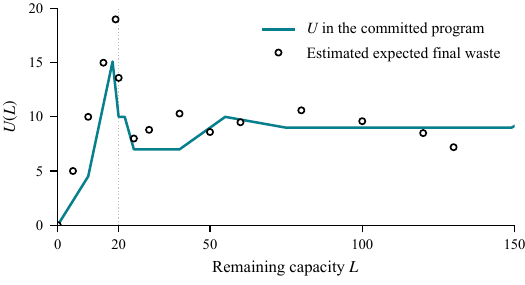}
    \caption{Estimated final waste and the optimized potential function, plotted against remaining capacity $L$, for the first commit of island 0 in GLM run 1 on OBP. Circles: the estimate from the agent's seventh analysis on self-generated sequences. Line: the table $U$ used by the committed program after coordinate descent. The dotted line marks the smallest item size, 20.}
    \label{fig:case}
\end{figure}

\subsection{A Search Trajectory}\label{sec:trajectory}
To show what the analysis contributes beyond the formal feedback, we trace one commit from its analyses to its hidden-set score: the first commit of island 0 in AgenticEvolve run 1 with GLM on OBP. After ten analyses, the agent submitted a new rule to replace the initial Best Fit rule.

The agent first examined where Best Fit wastes capacity. It generated 400 sequences from the stated distribution, checked that the excess of Best Fit on them was close to its formal value, and recorded the remaining capacity of each used bin at the end of each sequence. Bins left with less than 20 units, which no item can fill, accounted for 58\% of the total waste. The agent then estimated the expected final waste $U(L)$ of a bin with current remaining capacity $L$ under the rule it was testing (Fig.~\ref{fig:case}). Below 20 the estimate equals $L$ itself. Above 20 it falls quickly to between 7 and 11 units and remains in that range at the larger capacities observed.

The agent turned this estimate into a placement rule and tested it before committing. Placing an item of size $s$ into a bin with remaining capacity $L$ changes the expected waste of that bin from $U(L)$ to $U(L-s)$. The rule scores each feasible bin by $U(L)-U(L-s)$, adds a small Best Fit term to break ties, and selects the highest-scoring bin. The agent represented $U$ by 13 interpolation knots and adjusted their values by coordinate descent on its 400 sequences. The adjustment preserved the broadly flat shape above 20 and lowered the values below it (Fig.~\ref{fig:case}). On four sets of self-generated sequences the excess fell from 6.3 to 6.6\% under Best Fit to 5.4 to 5.7\% under the new rule. On the hidden set the committed program reduced the excess from 6.553\% to 5.392\%. Section S2-C lists the ten analyses with their results and gives the committed program.

\section{Symbolic Regression}\label{sec:sr}
The second experiment evaluates AgenticEvolve on symbolic regression, where the candidate program represents a mathematical expression fitted to data. Methods with analysis tools have access to the training data, so AgenticEvolve can fit and score a candidate during analysis using the same data as the evaluator. The in-domain and out-of-domain test sets are hidden and used only for reporting. We compare AgenticEvolve and No-analysis with LLM-SR~\cite{llmsr} and SR-Scientist~\cite{srscientist}, two LLM-based symbolic regression systems, and with PySR~\cite{pysr}, a genetic programming system, on twelve problems from LLM-SRBench~\cite{llmsrbench}.

\subsection{Protocol}
The twelve problems comprise all LSR-Synth problems that satisfy three criteria fixed before the runs: the complexity of the ground-truth expression lies within the interquartile range for its domain, the out-of-domain test split is informative, and removing any term from the ground truth and refitting the remaining terms gives a normalized mean squared error (NMSE) above $10^{-3}$ on the training data. The selection contains two chemistry problems (CRK21, CRK35), one biology problem (BPG18), six physics problems (PO0, PO1, PO7, PO10, PO16, PO26) and three material-science problems (MatSci6, MatSci9, MatSci11).

The four LLM-based methods use a common evaluator during search. A candidate is a Python function of the inputs and up to ten free constants. The evaluator traces the function into an expression tree and rejects trees with more than 30 nodes or with operators outside a fixed set. It fits the constants to the training data by least squares from a common starting point and returns the negative NMSE as the fitness. PySR retains its own search and constant fitting. For final selection, the evaluated candidates from each LLM-based run and the candidates on PySR's Pareto front undergo the same refitting and selection procedure. We construct the Pareto front of training error against node count, retain candidates whose error is at most $1.5$ times the minimum, and select the one with the highest PySR score~\cite{pysr}.

We assess whether each selected equation recovers the ground truth exactly or with negligible extra terms (Section S3-A). An equation is exact when it is algebraically equivalent to the ground truth up to reparameterization of the constants. Extra terms are negligible when their amplitude on the training data remains below $10^{-3}$ of the target's standard deviation. We also record whether an exact equation appeared anywhere in the run's candidate set. In the three material-science problems, the two inputs are coupled along $T = 273 + 500\varepsilon$ in every split, so the data cannot separate their effects. We therefore assess recovery against the ground truth reduced along this trajectory.

The four LLM-based methods follow the LLM-SRBench budget of up to 1000 model requests per problem. AgenticEvolve and No-analysis use three islands, each stopping at 333 commits, 333 requests or five consecutive turns without a tool call, whichever comes first. AgenticEvolve may run up to ten analyses per commit. LLM-SR draws 1000 samples per problem, and SR-Scientist runs up to 40 rounds of at most 25 model messages. Both retain their original search procedures but use the common evaluator (Section S1-C). This changes their constant fitting: on CRK21, LLM-SR's original BFGS routine leaves a substantially higher training error than least squares even when given the ground-truth expression. PySR uses the settings published by SR-Scientist, including a limit of 30 nodes. The four LLM-based methods use DeepSeek-V4.1-Flash with high reasoning effort, and costs are calculated using the prices in Section~\ref{sec:design}. To keep the five-method comparison within our computational budget, we run each method once per problem.

\subsection{Accuracy and Structure Recovery}
Table~\ref{tab:sr-summary} reports how many problems each method solves within the test tolerance of $10^{-3}$. AgenticEvolve meets the out-of-domain tolerance on all twelve problems, No-analysis and SR-Scientist on eight, LLM-SR on three and PySR on one. High predictive accuracy does not guarantee structural recovery. No-analysis on PO1 and SR-Scientist on CRK21 meet the tolerance with equations whose structure differs from the ground truth, as do AgenticEvolve, No-analysis and SR-Scientist on MatSci11. Table~\ref{tab:sr-structure} classifies the structure of each selected equation alongside its out-of-domain error, and Section S3-B lists the equations.

\begin{table}[t]
\centering
\caption{Symbolic regression results, totals over the twelve problems. Acc ID and Acc OOD count the problems whose selected equation is within tolerance $10^{-3}$ on the in-domain and out-of-domain test sets. For SR-Scientist, evaluations are the calls of its evaluator tool.}
\label{tab:sr-summary}
\footnotesize
\setlength{\tabcolsep}{3pt}
\begin{tabular}{lccrrr}
\toprule
Method & \begin{tabular}[b]{@{}c@{}}Acc\\ID\end{tabular} & \begin{tabular}[b]{@{}c@{}}Acc\\OOD\end{tabular} & Eval. & Requests & \begin{tabular}[b]{@{}c@{}}Cost\\(\$)\end{tabular} \\
\midrule
AgenticEvolve & 12 & 12 & 316 & 4939 & 6.54 \\
AgenticEvolve (No-analysis) & 8 & 8 & 3988 & 7809 & 12.12 \\
LLM-SR & 3 & 3 & 12000 & 12000 & 111.34 \\
SR-Scientist & 7 & 8 & 83 & 5496 & 6.59 \\
PySR & 1 & 1 & -- & -- & -- \\
\bottomrule
\end{tabular}
\end{table}

AgenticEvolve recovers the structure of all nine chemistry, biology and physics problems, six exactly and three with negligible extra terms. No-analysis and SR-Scientist each recover four of the nine, LLM-SR one and PySR one. On the material-science problems, No-analysis and SR-Scientist select the exact reduced forms of MatSci6 and MatSci9, while AgenticEvolve and LLM-SR select these forms with negligible extra terms. These polynomials in $\varepsilon$ describe the relationship along the data trajectory, leaving the independent temperature dependence undetermined. No method recovers the reduced form of MatSci11, whose Arrhenius factor varies by 1.2\% along the data trajectory and is not identifiable at the stated tolerance.

The last row of Table~\ref{tab:sr-structure} records whether an exact equation appeared anywhere in the search. AgenticEvolve's candidate sets contain exact equations for eleven problems, all but MatSci11. For five of them (CRK35, PO0, PO16, MatSci6, MatSci9), the final selection retains negligible extra terms even though an exact equation is available (Section S3-C). No-analysis and SR-Scientist find exact equations for five problems each, LLM-SR and PySR for one each.

\begin{table*}[t]
\centering
\caption{Structure of the selected equation on each problem ($\bullet$ exact, $\circ$ negligible extra terms, -- not recovered) and its normalized mean squared error on the out-of-domain test set, bold when within tolerance $10^{-3}$. The material-science problems are judged against the reduced truth.}
\label{tab:sr-structure}
\small
\begin{tabular}{lrrrrr}
\toprule
Problem & AgenticEvolve & \begin{tabular}[b]{@{}c@{}}AgenticEvolve\\(No-analysis)\end{tabular} & LLM-SR & SR-Scientist & PySR \\
\midrule
CRK21 & $\bullet$ \textbf{6.8e-15} & -- 2.0e-05 & -- 1.2e-05 & -- \textbf{5.6e-07} & -- 4.8e-05 \\
CRK35 & $\circ$ \textbf{3.0e-14} & $\bullet$ \textbf{2.9e-14} & -- 9.2e-05 & $\bullet$ \textbf{2.9e-14} & -- 3.8e-03 \\
BPG18 & $\bullet$ \textbf{1.7e-12} & $\bullet$ \textbf{1.7e-12} & $\circ$ \textbf{1.7e-12} & $\circ$ \textbf{2.2e-14} & -- 3.4e-03 \\
PO10 & $\bullet$ \textbf{1.4e-14} & -- 1.8e-02 & -- 2.0e-02 & $\bullet$ \textbf{1.4e-14} & -- 5.5e-04 \\
PO0 & $\circ$ \textbf{6.8e-13} & -- 7.2e-05 & -- 6.9e-04 & -- 1.2e-03 & -- 1.1e-02 \\
PO26 & $\bullet$ \textbf{1.0e-12} & $\circ$ \textbf{1.0e-12} & -- 1.5e-03 & -- 1.1e-02 & -- 3.2e-02 \\
PO16 & $\circ$ \textbf{1.3e-13} & -- 5.8e-06 & -- 1.7e-03 & $\circ$ \textbf{1.3e-13} & $\bullet$ \textbf{3.7e-10} \\
PO1 & $\bullet$ \textbf{9.6e-14} & -- \textbf{2.3e-09} & -- 2.6e-04 & -- 5.4e-03 & -- 6.9e-07 \\
PO7 & $\bullet$ \textbf{1.1e-14} & $\bullet$ \textbf{1.1e-14} & -- 8.2e-03 & -- 3.2e-02 & -- 5.8e-04 \\
\midrule
MatSci11 & -- \textbf{1.4e-12} & -- \textbf{4.0e-07} & -- 3.3e-04 & -- \textbf{1.4e-12} & -- 7.1e-02 \\
MatSci9 & $\circ$ \textbf{8.1e-13} & $\bullet$ \textbf{1.0e-12} & $\circ$ \textbf{8.1e-13} & $\bullet$ \textbf{1.0e-12} & -- 6.2e-04 \\
MatSci6 & $\circ$ \textbf{9.0e-13} & $\bullet$ \textbf{8.9e-13} & $\circ$ \textbf{9.3e-13} & $\bullet$ \textbf{1.1e-12} & -- 1.6e-01 \\
\midrule
Exact & 6 & 5 & 0 & 4 & 1 \\
Exact or negligible extras & 11 & 6 & 3 & 6 & 1 \\
Exact equation in candidate set & 11 & 5 & 1 & 5 & 1 \\
\bottomrule
\end{tabular}
\end{table*}

\subsection{Search Resources}
At comparable total cost, AgenticEvolve recovers the structure of more equations than SR-Scientist. The two methods spend 6.5 and 6.6 USD over the twelve problems (Table~\ref{tab:sr-summary}) and recover eleven and six equations exactly or with negligible extra terms. No-analysis spends 12.1 USD, roughly twice as much as AgenticEvolve, and LLM-SR spends 111 USD. Output tokens account for almost all of LLM-SR's cost: its 12000 samples produce 185 million output tokens, 99\% of them reasoning, compared with 7 to 13 million for each of the other three LLM-based methods (Section S3-D). Like OpenEvolve in Section~\ref{sec:heuristics}, LLM-SR has no execution tools, and DeepSeek reasons at length in each request.

The number of formal evaluations differs by two orders of magnitude between the methods and reflects how each one uses the evaluator. LLM-SR evaluates every sample it generates. No-analysis relies on formal feedback to guide its search, making 3988 commits in 7809 requests. AgenticEvolve makes 316 commits in 4939 requests. It experiments locally before committing, and 33 of its 36 islands stop after 2 to 21 commits when the agent makes no further tool calls. SR-Scientist calls the evaluator 83 times and its data analysis tool 5421 times. On each of PO1 and PO7, it evaluates only two distinct equations while making nearly a thousand analysis calls.

\subsection{Analysis Case}
BPG18 illustrates how analysis helps determine a constant that is difficult to fit. The ground truth is a logistic growth rate with a periodic forcing term,
\begin{equation}
    \dot P = P\,(a + bP + c\sin\omega t),
    \label{eq:bpg18}
\end{equation}
with $\omega \approx 0.77606$. For fixed $\omega$, fitting $a$, $b$ and $c$ is a linear least squares problem. When $\omega$ is also free, the common optimizer converges to an incorrect frequency from its default starting point. Every method that submitted this form with a free $\omega$ obtained an NMSE between 0.93 and 0.97. Each AgenticEvolve island used analysis to scan trial frequencies, fit the remaining coefficients at each frequency, and refine the frequency giving the lowest error. It then embedded the fitted frequency in the program as a literal. Two islands each recovered the exact structure in their first commit after ten analyses. The third initially submitted a free-frequency model and recovered the exact structure in its second commit after four further analyses.

The other methods reached the same structure by other routes, or not at all. No-analysis searched for the frequency through repeated commits, using the returned NMSE to guide each trial. Starting at commit 54, island 0 submitted the periodic form with a trial frequency encoded as a literal. At commit 107, $\omega = 0.77605$ met the exact-recovery criterion, and subsequent commits refined the frequency further. The other two islands built on island 0's commits. SR-Scientist used its data analysis tool to perform the same type of frequency scan as AgenticEvolve and submitted the exact structure at its second evaluation. Its final selected equation includes an additional correction for single-precision rounding of the stored time values, replacing $t$ with $t - \sin(\kappa(t-1))/\kappa$, where $\kappa \approx 532$. This equation is classified as the ground truth with negligible extra terms. LLM-SR first recovered the ground truth with negligible extra terms at its 209th sample. Its expression includes a phase parameter, and the common optimizer successfully fits the frequency in this parameterization. PySR selected a different structure, with an out-of-domain NMSE of $3.4\times10^{-3}$.

\section{Discussion}\label{sec:discussion}
The analysis tool lets the agent design its own procedure for the problem at hand. In the trajectory of Section~\ref{sec:trajectory}, the agent measured where Best Fit wastes capacity, estimated the expected final waste of a bin, built a placement rule from the estimate and optimized the rule locally before submitting it. On BPG18, it treated the frequency as a separate search problem because the common optimizer struggled to fit it. It scanned trial frequencies on the training data and submitted an expression with the fitted frequency fixed. In both cases, analysis provided the information and computation used to construct the candidate, which the agent tested locally before submitting it for formal evaluation.

What the analysis can establish depends on the data the task exposes. In symbolic regression, the agent can inspect the training data directly and design computations such as the frequency scan on BPG18. It can also fit and score a candidate on the same data as the evaluator. Formal evaluation therefore checks a candidate whose training performance the agent can already assess locally. In heuristic optimization, the evaluation instances are hidden. The agent generates its own instances from the stated distribution, and improvements on these instances must still be verified on the hidden set. The case in Section~\ref{sec:trajectory} illustrates this transfer, and the results in Section~\ref{sec:heuristics} show that AgenticEvolve's improvements hold on both hidden and unseen instances with every model.

Prompt caching makes the persistent conversation affordable. Each request resends the conversation, which grows with every analysis, so an AgenticEvolve run sends 24 to 164 times as many input tokens as an OpenEvolve run. With 99\% of these tokens served from cache, AgenticEvolve's mean cost ranges from 0.6 to 4.5 USD per run across models and tasks. Cached input accounts for about a third of this cost with DeepSeek and about four fifths with GLM and Qwen, whose cache-read prices are higher than DeepSeek's. Without execution tools, the model produces far more reasoning tokens. With DeepSeek, OpenEvolve generates 15 to 27 thousand output tokens per request, compared with about one thousand for AgenticEvolve, making OpenEvolve more expensive despite its shorter prompts. For the same reason, LLM-SR costs seventeen times as much as AgenticEvolve in symbolic regression. Persistent conversations also require management of the limited context window. In 15 of the 30 AgenticEvolve heuristic runs, at least one session reached the compaction threshold.

Execution inside the candidate program provides another means of computation that the evaluation protocol must account for. Four of the 90 final heuristic programs tune their parameters at import time within a wall-clock limit, making their scores sensitive to the load on the evaluating machine. Re-evaluation changed the score of one of them by less than $10^{-3}$ and left the other three unchanged. The feedback restriction in Section~\ref{sec:protocol} followed an earlier No-analysis run on BPG18 in which a candidate located the training data using an evaluator path exposed in a traceback. Only the No-analysis runs were repeated under the restriction. In the final runs, agents still attempted to access hidden data through analysis scripts and candidate programs. The audit found no successful access (Section S4). Both types of code run as ordinary host processes, so access restrictions are protocol-level and compliance is checked by audit. Enforcing file-access boundaries would require an operating-system sandbox with explicit access controls.

Agents used the archive and migration to inspect and build on other islands' programs, although the effect of these mechanisms was not measured in isolation. Every AgenticEvolve heuristic run included reads of programs from other islands, and about a quarter of the candidates offered through migration were read. Of the 215 commits preceded by a cross-island read, 169 improved the island's best score and 29 improved the run's best score (Section S2-D). In one OBP run with Qwen, the agent adopted a neighboring island's value table and extended its rule for opening a new bin, identifying the source in its commit description. These observations associate cross-island reads with subsequent improvements, but do not establish the effect of migration. That would require a comparison with migration disabled.

Each symbolic regression method ran once per problem, so run-to-run variation is not measured. All three models are Flash-tier models, and the benefit of analysis with stronger models remains untested. The task coverage is limited to two heuristic design problems and twelve benchmark equations. In the heuristic experiments, AgenticEvolve and No-analysis have the same formal evaluation limit. Whether No-analysis could close the quality gap with more evaluations at AgenticEvolve's total cost remains untested.

\section{Conclusion}
We presented AgenticEvolve, an evolutionary program search framework in which a persistent LLM agent conducts the search on each island and runs experiments between formal evaluations. On two heuristic design tasks, these experiments raised final program quality with all three models, and the improvement carried over to unseen instances. AgenticEvolve reached the quality of No-analysis (with experiments disabled) and OpenEvolve with a fraction of their formal evaluations. Prompt caching kept the cost of its persistent conversation within a few USD per run. In equation discovery, where the agent can inspect the data, AgenticEvolve recovered more benchmark equations than No-analysis, LLM-SR, SR-Scientist and PySR. In both settings, the agent designed experiments to guide candidate construction, using self-generated instances for heuristic optimization and the available data for equation discovery.

We adopted the population, the archive, the islands and migration from evolutionary computation, where these mechanisms were developed for variation operators specified in advance. A persistent agent reads the archive, chooses which program to build on and runs its own experiments before submitting a candidate. It remains open how much of this structure an agent-driven search needs, and how the search should be divided between the agent and the evolutionary mechanisms.

\section*{Code and Data Availability}
The source code is available at \url{https://github.com/pengzy-cosmo/AgenticEvolve}. Source snapshots for all versions used in the experiments, experimental run records, and evaluation-access audits are archived at \url{https://doi.org/10.5281/zenodo.23561528}. The LLM-SRBench data are available from \url{https://huggingface.co/datasets/nnheui/llm-srbench} under the dataset's access conditions and are not redistributed in the archive.

\section*{Acknowledgment}
The authors acknowledge the use of generative AI tools, including Claude and ChatGPT, for language editing of the manuscript and the supplementary material. The authors reviewed and revised all edits and take full responsibility for the content.

\bibliographystyle{IEEEtran}
\bibliography{refs}

\onecolumn
\setcounter{section}{0}
\setcounter{subsection}{0}
\setcounter{table}{0}
\setcounter{figure}{0}
\setcounter{equation}{0}
\renewcommand{\thesection}{S\arabic{section}}
\renewcommand{\thetable}{S\arabic{table}}
\renewcommand{\thefigure}{S\arabic{figure}}
\renewcommand{\theequation}{S\arabic{equation}}
\begin{center}
    {\Large Supplementary Material}
\end{center}
\section{Experimental Settings}\label{supp:sec:settings}

\subsection{Model Requests}
AgenticEvolve and OpenEvolve use the same generation parameters for each model. AgenticEvolve uses the providers' default temperature of 1.0, while OpenEvolve sets the temperature explicitly to the same value. DeepSeek ignores the temperature in thinking mode. The output limit is set to each model's maximum: 384000 tokens for DeepSeek and 131072 tokens for GLM and Qwen, including reasoning tokens. Table~\ref{tab:models} of the paper gives the reasoning effort.

Request handling differs between the two systems. AgenticEvolve includes tool schemas in its requests and streams each response, with a timeout of 300 seconds between streamed chunks. OpenEvolve uses non-streaming requests and includes a random seed. Its timeout applies to the entire request and is set to 1200 seconds to accommodate DeepSeek's reasoning, which takes about five minutes on OpenEvolve's prompt.

AgenticEvolve limits the conversation context to 500000 tokens. When the context exceeds 70\% of this limit, the next commit feedback asks the agent to end its turn, after which the framework compacts the conversation (Section~\ref{sec:island}). At 90\% the framework cancels the turn immediately. Compaction occurred 25 times across 15 of the 30 AgenticEvolve heuristic runs. No-analysis never reached the threshold.

\subsection{OpenEvolve Settings}
OpenEvolve~0.3.2 runs with its default settings except those in Table~\ref{tab:openevolve}, which were fixed before the formal runs.

\begin{table}[h]
    \centering
    \caption{OpenEvolve settings that differ from the defaults of version 0.3.2. The evaluator timeout is given for OBP and DJSS.}
    \label{tab:openevolve}
    \small
    \begin{tabular}{@{}l>{\raggedright\arraybackslash}p{0.11\textwidth}>{\raggedright\arraybackslash}p{0.1\textwidth}>{\raggedright\arraybackslash}p{0.36\textwidth}@{}}
        \toprule
        Setting & Default & Used & Reason \\
        \midrule
        \texttt{num\_islands} & 5 & 3 & Same as AgenticEvolve. \\
        \texttt{feature\_dimensions} & code length, diversity & task features & Same features as AgenticEvolve. \\
        \texttt{migration\_interval} & 50 & 10 & Each island advances about 60 generations in a run, so the default would trigger migration at most once. \\
        \texttt{population\_size} & 1000 & 50 & A run produces 180 programs and would never fill the default population. Most example configurations distributed with OpenEvolve use 40 to 100. \\
        \texttt{temperature} & 0.7 & 1.0 & Same as AgenticEvolve. \\
        \texttt{max\_tokens} & 4096 & model maximum & Same as AgenticEvolve. \\
        LLM \texttt{timeout} & 60~s & 1200~s & DeepSeek reasons for about five minutes per request. \\
        LLM \texttt{retries} & 3 & 2 & Three attempts, matching AgenticEvolve when the settings were fixed. AgenticEvolve later increased this limit to nine (Section~\ref{supp:sec:versions}). \\
        \texttt{suggest\_simplification\_after\_chars} & 500 & disabled & Code length is not part of the fitness, and the suggestion would steer the search toward shorter code. \\
        evaluator \texttt{timeout} & 300~s & 90 / 180~s & The shared executor enforces the actual limits of 60 and 150~s. \\
        evaluator \texttt{max\_retries} & 3 & 0 & Each program is evaluated once, as in AgenticEvolve. \\
        \texttt{cascade\_evaluation} & true & false & The shared evaluators have a single stage. \\
        \bottomrule
    \end{tabular}
\end{table}

OpenEvolve receives the task description as its system message. We replace its model client to send the output limit through the API field that includes reasoning tokens and to record token usage for every request. All other settings retain their defaults, including diff-based generation, the code-length limit of 10000 characters, parent sampling and the rest of the prompt template.

We retain three differences between OpenEvolve and AgenticEvolve as part of the system comparison. OpenEvolve rescales each feature using the minimum and maximum observed so far before assigning it to a bin, whereas AgenticEvolve uses the evaluator's fixed normalization. An OpenEvolve iteration whose response contains no valid diff consumes a model request without an evaluation. We therefore count formal evaluations in the evaluator adapter rather than use the iteration count. OpenEvolve's prompt shows metrics to four decimal places, whereas AgenticEvolve shows them unrounded.

\subsection{Symbolic-Regression Baselines}
We use the LLM-SR implementation from the LLM-SRBench repository at commit 6fdbe409. Its search is unchanged, with ten islands, two example programs and four samples per prompt, and a time limit of 30 seconds per evaluation. The shared evaluator replaces the original BFGS fitting procedure, and the prompt states the shared evaluation rules. Our model client replaces the repository's client, which caps every response at 1024 tokens and records no token usage.

We use the SR-Scientist inference code at commit 00a44a1. Its search is unchanged, with 40 rounds of at most 25 model messages, an error target that tightens from round to round, and the three best candidates carried into the next round. The evaluator tool and the re-evaluation at the end of each round both use the shared evaluator in place of the original BFGS fitting procedure. The system prompt and tool description state the shared evaluation rules. The data analysis tool executes the model's code in a local subprocess that receives the training data on standard input, replacing the original remote sandbox service. Our model client replaces the original client, which caps every response at 8192 tokens and records no token usage. The replacement clients for both LLM-SR and SR-Scientist use the model's maximum output length and high reasoning effort, matching AgenticEvolve.

PySR uses the settings published with SR-Scientist for this benchmark: 125 iterations of 550 cycles, 15 populations of 33 members, a maximum of 30 nodes, and the operators $+$, $-$, $\times$, $\div$, power, $\exp$, $\log$, $\sqrt{\cdot}$, $\sin$ and $\cos$ with their nesting constraints. We omit the published cap of 100000 candidates per problem because it stops the search within the first iteration under these settings.

\subsection{System Prompt}
Every island of AgenticEvolve receives the following system prompt, with the fields in braces filled from the task configuration.

{\small
\begin{verbatim}
You are the engine of an evolutionary program search: you improve a Python
program against a formal evaluator.

Task:
{task description}

Feedback:
- The evaluator returns the metrics described in the task; the fitness is
  {fitness}, higher is better.
- An archive keeps every commit and, per island, a MAP-Elites grid over the
  feature fields {features} (10 bins per axis,
  each field normalized to [0, 1]) holding the fittest commit of each cell.
- You run one of several islands searching in parallel. Each commit returns
  the commit's metrics, the island's best commit, its MAP-Elites, and
  periodically migration candidates: top elites of the neighboring islands.

Files, in memory and accessible only through your tools:
- program.py: the evolving candidate; only this file is formally evaluated.
- analysis.py: your workspace for investigating hypotheses; it may `import program`.
Every execution runs in a fresh temporary directory holding program.py and analysis.py:
only stdout comes back, do not touch files outside it, and state persists
across executions only through program.py and analysis.py.

Tools:
- read / write / edit: operate on program.py and analysis.py.
- run_analysis: execute the current analysis.py and return its output, up to
  10 runs per commit cycle.
- commit: run the formal evaluator on program.py, append the program and result
  to the archive, and return the feedback above. Formal metrics are available
  only through commit. A failed evaluation is recorded and consumes budget like
  any other commit.
- read_commit: return the full program.py of a recorded commit; write it to
  program.py to continue from it.

Experiment loop, repeated until the framework tells you to stop:
1. Form a hypothesis from the current program and the feedback so far.
2. Modify program.py accordingly.
3. Commit the experiment and inspect its formal feedback.
4. Use the result, the MAP-Elites, and the migration candidates to choose the
   next idea.

At any point, use analysis.py and run_analysis to explore the data, check an
idea, or diagnose a result.

You decide what to investigate, how much analysis to perform, and what to
change. When a message tells you to end the turn, comply immediately.
\end{verbatim}
}

No-analysis removes the line describing \texttt{analysis.py}, the tool \texttt{run\_analysis} and the paragraph beginning ``At any point''. Its file list names only \texttt{program.py}, and its closing instruction reads ``You decide what to investigate and what to change. When a message tells you to end the turn, comply immediately.'' In the symbolic-regression runs of AgenticEvolve, the prompt also states that the temporary directory contains the task data in \texttt{data/}. Version 0.2.1 adds the following to the description of \texttt{commit}: ``its feedback is the exception type and line, plus the message when the evaluator itself rejected the program.'' The prompt is otherwise identical in all versions.

The first message of an island contains the initial program, its evaluation and the budget. After each turn the framework sends the message ``Continue.'' When the context nears its limit, the next commit feedback ends with ``[context limit] Your context is nearly full. State your current hypothesis in one or two sentences, then end this turn immediately without calling any more tools.'' The agent then receives the following request for a handoff summary. A new conversation with the same system prompt starts from this summary and the island's current state.

{\small
\begin{verbatim}
Ignore the running contract; do not call tools. Your context is being reset. A
successor will continue from the current program.py and the archive state, but
without this conversation. Write a concise handoff covering exactly:
1. Approaches already tried that failed, and why each failed.
2. Facts about the problem and data established by your experiments.
3. Your current hypothesis.
4. The next step you would take.
\end{verbatim}
}

When the budget of an island is exhausted, the last feedback ends with ``[budget exhausted] The run budget is exhausted. End this turn immediately without calling any more tools.''

\subsection{Framework Versions}\label{supp:sec:versions}
The heuristic runs used three successive framework versions, 0.1.0 to 0.1.2. Version 0.1.1 increased the maximum number of attempts per model request from three to nine, and version 0.1.2 added the scripts for the generalization tests. The search, evaluators and budgets were unchanged across these versions. In the symbolic-regression experiment, AgenticEvolve, LLM-SR, SR-Scientist and PySR ran on version 0.2.0 and No-analysis ran on version 0.2.1 (Section~\ref{supp:sec:audits}). Each run records the framework commit used.

\section{Heuristic Design Results}\label{supp:sec:heuristics}

\subsection{Final Quality and Generalization}
Table~\ref{tab:sets} reports final quality on each evaluation set, and Table~\ref{tab:differences} reports pairwise differences with confidence intervals and $p$ values. The statistics use five independent runs per method, model and task. Hidden-set values are taken from the run records, as in Table~\ref{tab:heuristics} of the paper. The final programs were subsequently re-evaluated on all sets. The re-evaluated hidden-set fitness agrees with the recorded value to six decimal places for every program except one DeepSeek No-analysis program on OBP. That program tunes its parameters at import time within a wall-clock limit, and its fitness changed by less than $10^{-3}$ (Section~\ref{sec:discussion} of the paper).


\begin{table}[h]
\centering
\caption{Final quality on every evaluation set, mean $\pm$ SD over five runs (lower is better). OBP quality is the excess over the $L_2$ bound in percent, and DJSS quality is the tardiness relative to rule R1. Hidden-set values are the scores recorded during the runs. Values on OR1 and on fresh instances come from a separate evaluation of the final programs.}
\label{tab:sets}
\small
\begin{tabular}{@{}llrrrrr@{}}
\toprule
 & & \multicolumn{3}{c}{Online bin packing} & \multicolumn{2}{c}{Dynamic job shop} \\
\cmidrule(lr){3-5}\cmidrule(l){6-7}
Model & Method & Hidden & OR1 & Fresh & Hidden & Fresh \\
\midrule
DeepSeek & AgenticEvolve & 3.42 $\pm$ 0.25 & 2.92 $\pm$ 0.48 & 3.72 $\pm$ 0.32 & 0.840 $\pm$ 0.063 & 0.846 $\pm$ 0.072 \\
 & AgenticEvolve (No-analysis) & 5.54 $\pm$ 1.05 & 5.12 $\pm$ 0.92 & 5.69 $\pm$ 0.95 & 0.970 $\pm$ 0.028 & 0.974 $\pm$ 0.025 \\
 & OpenEvolve & 5.74 $\pm$ 0.41 & 5.12 $\pm$ 0.42 & 5.85 $\pm$ 0.42 & 0.911 $\pm$ 0.041 & 0.923 $\pm$ 0.025 \\
\addlinespace[2pt]
GLM & AgenticEvolve & 4.46 $\pm$ 0.80 & 4.14 $\pm$ 0.54 & 4.72 $\pm$ 0.74 & 0.847 $\pm$ 0.039 & 0.863 $\pm$ 0.039 \\
 & AgenticEvolve (No-analysis) & 6.30 $\pm$ 0.19 & 5.61 $\pm$ 0.26 & 6.37 $\pm$ 0.16 & 0.928 $\pm$ 0.074 & 0.952 $\pm$ 0.069 \\
 & OpenEvolve & 5.54 $\pm$ 0.81 & 4.95 $\pm$ 0.57 & 5.72 $\pm$ 0.66 & 0.931 $\pm$ 0.054 & 0.944 $\pm$ 0.064 \\
\addlinespace[2pt]
Qwen & AgenticEvolve & 3.70 $\pm$ 0.52 & 3.10 $\pm$ 0.39 & 3.98 $\pm$ 0.61 & 0.820 $\pm$ 0.020 & 0.841 $\pm$ 0.027 \\
 & AgenticEvolve (No-analysis) & 5.98 $\pm$ 0.24 & 5.32 $\pm$ 0.32 & 6.08 $\pm$ 0.20 & 0.981 $\pm$ 0.045 & 0.978 $\pm$ 0.053 \\
 & OpenEvolve & 6.13 $\pm$ 0.10 & 5.50 $\pm$ 0.24 & 6.25 $\pm$ 0.14 & 0.978 $\pm$ 0.069 & 0.999 $\pm$ 0.057 \\
\midrule
\multicolumn{2}{@{}l}{Best Fit} & 6.55 & 5.81 & 6.39 & -- & -- \\
\multicolumn{2}{@{}l}{FunSearch heuristic} & 6.10 & 5.30 & 6.21 & -- & -- \\
\multicolumn{2}{@{}l}{Rule R1} & -- & -- & -- & 1.000 & 1.000 \\
\bottomrule
\end{tabular}
\end{table}

\begin{table}[h]
\centering
\caption{Differences in mean final quality between methods, with bootstrap 95\% confidence intervals and two-sided Mann--Whitney $p$ values over five runs per method. Each difference is the first method minus the second, so a negative value favors the first method. Units are those of Table~\ref{tab:sets}. The $p$ values are not corrected for multiple comparisons. $^{a}$Asymptotic approximation because of ties. All other $p$ values are exact.}
\label{tab:differences}
\small
\setlength{\tabcolsep}{4pt}
\begin{tabular}{@{}lllrrrrrr@{}}
\toprule
 & & & \multicolumn{2}{c}{AgenticEvolve $-$ No-analysis} & \multicolumn{2}{c}{AgenticEvolve $-$ OpenEvolve} & \multicolumn{2}{c}{No-analysis $-$ OpenEvolve} \\
\cmidrule(lr){4-5}\cmidrule(lr){6-7}\cmidrule(l){8-9}
Model & Task & Set & Difference & $p$ & Difference & $p$ & Difference & $p$ \\
\midrule
DeepSeek & OBP & Hidden & $-$2.12 [$-$2.93, $-$1.24] & 0.008 & $-$2.32 [$-$2.73, $-$1.98] & 0.008 & $-$0.20 [$-$1.14, 0.66] & 0.917$^{a}$ \\
 &  & OR1 & $-$2.20 [$-$3.00, $-$1.37] & 0.008 & $-$2.20 [$-$2.71, $-$1.69] & 0.012$^{a}$ & 0.00 [$-$0.84, 0.75] & 0.916$^{a}$ \\
 &  & Fresh & $-$1.97 [$-$2.73, $-$1.16] & 0.008 & $-$2.13 [$-$2.58, $-$1.75] & 0.008 & $-$0.16 [$-$1.01, 0.62] & 1.000 \\
 & DJSS & Hidden & $-$0.130 [$-$0.175, $-$0.072] & 0.016 & $-$0.071 [$-$0.123, $-$0.006] & 0.095 & 0.059 [0.022, 0.100] & 0.056 \\
 &  & Fresh & $-$0.128 [$-$0.176, $-$0.061] & 0.032 & $-$0.077 [$-$0.124, $-$0.009] & 0.151 & 0.051 [0.025, 0.079] & 0.032 \\
\addlinespace[3pt]
GLM & OBP & Hidden & $-$1.84 [$-$2.47, $-$1.19] & 0.008 & $-$1.07 [$-$1.93, $-$0.17] & 0.095 & 0.76 [0.14, 1.43] & 0.056 \\
 &  & OR1 & $-$1.47 [$-$1.92, $-$0.98] & 0.012$^{a}$ & $-$0.82 [$-$1.43, $-$0.18] & 0.075$^{a}$ & 0.65 [0.18, 1.14] & 0.059$^{a}$ \\
 &  & Fresh & $-$1.65 [$-$2.23, $-$1.05] & 0.008 & $-$1.00 [$-$1.75, $-$0.22] & 0.095 & 0.65 [0.15, 1.21] & 0.032 \\
 & DJSS & Hidden & $-$0.081 [$-$0.146, $-$0.014] & 0.222 & $-$0.084 [$-$0.138, $-$0.035] & 0.016 & $-$0.003 [$-$0.078, 0.066] & 1.000 \\
 &  & Fresh & $-$0.089 [$-$0.149, $-$0.023] & 0.032 & $-$0.080 [$-$0.143, $-$0.025] & 0.056 & 0.008 [$-$0.070, 0.076] & 1.000 \\
\addlinespace[3pt]
Qwen & OBP & Hidden & $-$2.28 [$-$2.66, $-$1.78] & 0.012$^{a}$ & $-$2.43 [$-$2.76, $-$1.95] & 0.012$^{a}$ & $-$0.15 [$-$0.36, 0.05] & 0.402$^{a}$ \\
 &  & OR1 & $-$2.22 [$-$2.57, $-$1.79] & 0.012$^{a}$ & $-$2.41 [$-$2.73, $-$2.02] & 0.012$^{a}$ & $-$0.18 [$-$0.51, 0.10] & 0.670$^{a}$ \\
 &  & Fresh & $-$2.10 [$-$2.49, $-$1.53] & 0.008 & $-$2.27 [$-$2.64, $-$1.70] & 0.012$^{a}$ & $-$0.17 [$-$0.38, 0.01] & 0.091$^{a}$ \\
 & DJSS & Hidden & $-$0.161 [$-$0.202, $-$0.125] & 0.008 & $-$0.158 [$-$0.217, $-$0.106] & 0.008 & 0.003 [$-$0.064, 0.066] & 1.000 \\
 &  & Fresh & $-$0.137 [$-$0.186, $-$0.093] & 0.008 & $-$0.157 [$-$0.205, $-$0.105] & 0.008 & $-$0.021 [$-$0.079, 0.044] & 0.548 \\
\bottomrule
\end{tabular}
\end{table}

\FloatBarrier
\subsection{Budgets to Reach OpenEvolve}
Table~\ref{tab:reach} reports the budgets at which AgenticEvolve reaches OpenEvolve's mean final quality over five runs. For each resource axis, we average the best-so-far quality of the five AgenticEvolve runs at each budget and report the first budget at which this mean reaches the target. On the evaluation axis, this average is defined at each of the 91 evaluations completed by all five runs. On the cost axis, completed runs retain their final quality at larger budgets, as in Fig.~\ref{fig:cost} of the paper.

\begin{table}[h]
\centering
\caption{Formal evaluations and estimated cost at which the mean best-so-far curve of AgenticEvolve in Figs.~\ref{fig:evaluations} and \ref{fig:cost} of the paper first reaches the mean final quality of OpenEvolve. The two budgets are read from the two curves separately and do not refer to the same point of the search. The last column is the mean cost of a complete OpenEvolve run.}
\label{tab:reach}
\small
\begin{tabular}{@{}llrrrr@{}}
\toprule
 & & OpenEvolve & \multicolumn{2}{c}{AgenticEvolve} & OpenEvolve \\
\cmidrule(lr){4-5}
Model & Task & final quality & Evaluations & Cost (\$) & cost (\$) \\
\midrule
DeepSeek & OBP & 5.74 & 7 & 0.15 & 3.06 \\
GLM & OBP & 5.54 & 17 & 0.96 & 0.69 \\
Qwen & OBP & 6.13 & 3 & 0.08 & 1.10 \\
DeepSeek & DJSS & 0.911 & 19 & 0.19 & 1.84 \\
GLM & DJSS & 0.931 & 26 & 0.20 & 0.48 \\
Qwen & DJSS & 0.978 & 8 & 0.12 & 0.64 \\
\bottomrule
\end{tabular}
\end{table}

\FloatBarrier
\subsection{A Search Trajectory}
Table~\ref{tab:case} traces the ten analyses preceding the commit examined in Section~\ref{sec:trajectory} of the paper. The agent generated sequences from the distribution stated in the task: 120 items per sequence, integer item sizes between 20 and 100, and bin capacity 150. Its simulator computed excess over the $L_2$ bound in the same way as the evaluator. The table expresses the printed fractional values as percentages.

\begin{table}[h]
    \centering
    \caption{The ten analyses before the first commit of island~0 in AgenticEvolve run~1 with GLM on OBP. Excess is over the $L_2$ bound on the agent's own sequences.}
    \label{tab:case}
    \small
    \begin{tabular}{@{}r>{\raggedright\arraybackslash}p{0.5\textwidth}>{\raggedright\arraybackslash}p{0.4\textwidth}@{}}
        \toprule
        & Analysis & Result \\
        \midrule
        1 & Simulated Best Fit and Worst Fit on 400 sequences and recorded each bin's final remaining capacity. & Best Fit excess 6.46\%, compared with 6.55\% in the formal evaluation. Bins with less than 20 units remaining account for 58\% of the waste. \\
        2 & Scored bins using a piecewise-linear table indexed by the capacity remaining after placement. Starting from Best Fit, adjusted 13 knot values by coordinate descent on 200 sequences. & Excess 5.99\% on all 400 sequences. \\
        3 & Repeated the descent with 20 knots on 400 sequences. & Exceeded the 60-second analysis limit. \\
        4 & Vectorized the simulator, repeated the 20-knot descent and tested the result on 400 sequences from a new seed. & 5.99\% on the training sequences and 6.19\% on the new ones, where Best Fit gives 6.59\%. The descent made no further progress. \\
        5 & Repeated the descent over all 131 table entries. & Exceeded the 60-second limit. \\
        6 & Counted the frequencies of the remaining capacities that would result from placing an item in each candidate bin, then repeated the descent on the eight most frequent values. & No change. \\
        7 & Estimated the expected final waste $U(L)$ of a bin with remaining capacity $L$ under the rule from analysis~4. Tested a new rule that maximizes $U(L)-U(L-s)$ for an item of size $s$. & $U(L)=L$ below 20, 13.6 at 20, and between 7 and 11 above 20 (Fig.~\ref{fig:case} of the paper). The rule gave 6.10\% excess because the estimate lacked observations of empty bins and assigned them a value of zero. \\
        8 & Searched a grid of value functions that equal $L$ below a threshold and are linear above it, combined with a small Best Fit term. & 5.66\% on the training sequences and 5.76\% on the new ones, with the threshold at 22 and a constant value of 11 above it. \\
        9 & Ran coordinate descent on the values of $U$ at 13 knots and on the weight of the Best Fit term, starting from a setting that reproduced Best Fit. & 5.41\% on the training sequences and 5.64\% on the new ones. \\
        10 & Compared Best Fit, the result of analysis~9 and a smoothed version of it on four sets of 400 sequences. & Best Fit 6.32 to 6.60\%, analysis~9 5.38 to 5.72\% and the smoothed version 5.78 to 5.95\%. \\
        \bottomrule
    \end{tabular}
\end{table}

The committed program, listed below, uses the knot values printed by analysis~9. Its hidden-set excess is 5.392\%, compared with 6.553\% for Best Fit. Although the docstring and commit description attribute $U$ to policy iteration, the committed values were obtained by coordinate descent in analysis~9.

{\small
\begin{verbatim}
"""Online bin packing: score each feasible bin for the arriving item.

Strategy: value-difference policy. Let U(L) be the expected final wasted
capacity of an open bin whose current remaining capacity is L (estimated
empirically under the evolving policy via policy iteration). Placing the
item into a bin changes its expected future waste from U(L) to U(L - item),
so we choose the bin that destroys the most expected waste, i.e. maximizes
U(L) - U(L - item), with a tiny best-fit tiebreak.

The learned U has a key structural feature: leftovers below ~20 can never
be filled (all items are >= 20), so they are dead weight (U(L) ~= L),
while any leftover >= 20 still expects to be filled down to ~7-10 waste.
"""

import numpy as np

_KNOTS = [0, 10, 18, 20, 22, 25, 30, 40, 55, 75, 100, 130, 150]
_UVALS = [0.0, 4.5, 15.1, 10.0, 10.0, 7.0, 7.0, 7.0, 10.0, 9.0, 9.0, 9.0, 9.0]

_U = np.interp(np.arange(151), _KNOTS, _UVALS)
# Expected waste of a fresh (empty) bin, averaged over the item distribution.
_U[150] = _U[50:131].mean()

_EPS = 0.005  # best-fit tiebreak strength

def priority(item: int, bins: np.ndarray) -> np.ndarray:
    """Returns the priority with which to add item to each feasible bin.

    Args:
        item: size of the arriving item.
        bins: remaining capacities of the feasible bins (each fits the item).

    Returns:
        Array of the same size as bins with the priority score of each bin.
    """
    r = bins - item
    return _U[bins] - _U[r] - _EPS * r
\end{verbatim}
}

\FloatBarrier
\subsection{Cross-Island Reads}
Table~\ref{tab:migration} summarizes cross-island reads in the 30 AgenticEvolve heuristic runs, supporting the observations in Section~\ref{sec:discussion} of the paper. For each receiving island, a migration candidate is counted as offered when it first appears in the migration feedback, provided the island commits at least once afterwards. It is counted as read if the agent subsequently retrieves its program with \texttt{read\_commit}. Each cross-island read is associated with the next commit, allowing for intervening analyses and edits. A commit improves the island's or run's best score if it exceeds the corresponding best score before that commit, including the initial program's score.

The column ``Offered candidates read'' gives the number of offered candidates subsequently read out of the total offered. The other three columns count commits following any cross-island read, including reads of programs absent from the island's recorded migration feedback. There were 25 such reads, which contribute only to the commit-based statistics. Of the 215 commits following a cross-island read, 36 contain code identical to a program the agent read, and one more differs only in whitespace.

\begin{table}[h]
    \centering
    \caption{Cross-island reads in the AgenticEvolve heuristic runs, five runs per row.}
    \label{tab:migration}
    \small
    \begin{tabular}{@{}llrrrr@{}}
        \toprule
        Model & Task & Offered candidates read & Commits after a read & Improved island best & Improved run best \\
        \midrule
        DeepSeek & OBP & 53 / 154 & 42 & 34 & 6 \\
        DeepSeek & DJSS & 39 / 119 & 34 & 28 & 4 \\
        GLM & OBP & 42 / 153 & 38 & 29 & 6 \\
        GLM & DJSS & 28 / 142 & 25 & 18 & 5 \\
        Qwen & OBP & 42 / 174 & 41 & 37 & 4 \\
        Qwen & DJSS & 38 / 167 & 35 & 23 & 4 \\
        \midrule
        Total & & 242 / 909 & 215 & 169 & 29 \\
        \bottomrule
    \end{tabular}
\end{table}

\FloatBarrier
\section{Symbolic Regression}\label{supp:sec:sr}

\subsection{Structure Assessment}
The structure assessment in Section~\ref{sec:sr} assigns each selected equation to one of three tiers. An equation is exact when it equals the ground truth up to reparameterization of the constants. Integer exponents belong to the structure, so $x$ and $x^3$ are different terms. Every other numeric constant of the ground truth counts as free. This includes the 1 in $\log(kt+1)$ and $\log(|v|+1)$, the exponent $1/3$ in PO7 and the frequency and phase of each sinusoidal term. Repeated occurrences of a constant, such as $\omega_0^2$ in PO10, may be fitted separately.

An equation has negligible extra terms when it contains the ground truth and removing each additional term changes the output by less than $10^{-3}$ of the target's standard deviation on the training data. The same criterion applies to a term inside a product or a function. A missing term of the ground truth is never negligible. All other equations are not recovered.

Both recovered tiers also require an absolute error below $10^{-3}$ of the target's standard deviation at every training point. This condition must hold for the full equation and for the equation with any extra terms removed. An equation with the correct structure but constants that fail this accuracy requirement is classified as not recovered.

A single checking script applies these rules to every candidate across the 60 runs. It substitutes the fitted constants, expands the expression and groups terms by their dependence on the variables. Each group must match a ground-truth term. Constants close to an integer or a simple fraction are first rounded, provided the resulting change in output stays below the threshold. If the groups still do not match, the script removes terms one at a time under the same output-change criterion and repeats the comparison.

In the three material-science problems, strain and temperature follow $T = 273 + 500\varepsilon$ in every data split. Along this line the ground truth becomes a function of $\varepsilon$ alone, which Table~\ref{tab:sr-equations} lists as the reduced truth. The reduced truths of MatSci6 and MatSci9 are polynomials in $\varepsilon$. Recovering them does not recover the temperature dependence of the original laws. The reduced truth of MatSci11 retains the Arrhenius factor, and no method recovered it.

\subsection{Selected Equations}
Table~\ref{tab:sr-equations} lists the selected equation from each run and its classification, as reported in Table~\ref{tab:sr-structure} of the paper.

{\small
    \begin{longtable}{@{}l r c >{\raggedright\arraybackslash\hangindent=1.5em\hangafter=1}p{0.6\textwidth}@{}}
\caption{Selected equation of every method on every problem, with its number of nodes and its structure ($\bullet$ exact, $\circ$ negligible extra terms, -- not recovered). Constants are rounded to four significant digits for display. The assessment uses the full-precision values. Equations of the material-science problems are judged against the reduced truth after substituting $T = 273 + 500\varepsilon$.}\label{tab:sr-equations} \\
\toprule
Method & Nodes & Structure & Equation \\
\midrule
\endfirsthead
\toprule
Method & Nodes & Structure & Equation \\
\midrule
\endhead
\bottomrule
\endlastfoot
\multicolumn{4}{@{}l}{\textbf{CRK21} (chemistry), target $\dot A$} \\*
Ground truth & & & $-k_1 A e^{-k_2 t} + k_3 A \log(k_4 t + 1)$ \\*[3pt]
AgenticEvolve & 17 & $\bullet$ & $A (0.08820 \log{(t + 2.283 )} - 0.07279 - 0.9112 e^{- 0.08901 t})$ \\*[3pt]
AgenticEvolve (No-analysis) & 30 & -- & $A (t (t (t (2.600 \cdot 10^{-6} - 1.696 \cdot 10^{-8} t) - 0.0001558) + 0.004786) - 0.04734) (\sqrt{A} - 0.1362 t + 18.25) - 6.356 \cdot 10^{-5}$ \\*[3pt]
LLM-SR & 26 & -- & $0.2869 A^{1.014} - \frac{3.122 A^{1.015} e^{- 0.08080 t}}{A^{0.002530} - 0.02826 A + 1.633}$ \\*[3pt]
SR-Scientist & 24 & -- & $A (0.3531 - 0.02998 e^{- 0.6502 t} - 0.9788 e^{- 0.09575 t} - 0.2555 e^{- 0.02338 t})$ \\*[3pt]
PySR & 24 & -- & $0.008062 t (- 0.006810 A t + A) - 0.9108 e^{\frac{t}{- 0.05985 t - 0.9673}} - 0.0002335$ \\
\midrule
\multicolumn{4}{@{}l}{\textbf{CRK35} (chemistry), target $\dot A$} \\*
Ground truth & & & $-k_1 A + k_2 A \log(k_3 t + 1)$ \\*[3pt]
AgenticEvolve & 16 & $\circ$ & $- 6.284 \cdot 10^{-11} A^{2} + A (0.06226 \log{(t + 1.061 )} - 0.1400)$ \\*[3pt]
AgenticEvolve (No-analysis) & 10 & $\bullet$ & $A (0.06226 \log{(t + 1.061 )} - 0.1400)$ \\*[3pt]
LLM-SR & 26 & -- & $0.09319 A^{1.069} - 0.03329 + \frac{(- 0.7000 A^{3.635} - 0.3507) e^{- 0.1569 t}}{A^{1.824} + 4.375}$ \\*[3pt]
SR-Scientist & 10 & $\bullet$ & $A (0.06226 \log{(t + 1.061 )} - 0.1400)$ \\*[3pt]
PySR & 21 & -- & $0.1194 A - 0.1194 ({0.5452}^{\frac{\sqrt{t + 0.5089} - e^{{0.09630}^{A}}}{A}} + 0.5202)^{1.273}$ \\
\midrule
\multicolumn{4}{@{}l}{\textbf{BPG18} (biology), target $\dot P$} \\*
Ground truth & & & $r \left(1 - \frac{P}{K}\right) P + r P + c P \sin(\omega t)$ \\*[3pt]
AgenticEvolve & 14 & $\bullet$ & $P (- 0.008724 P + 0.4771 \sin{(0.7761 t )} + 0.8909)$ \\*[3pt]
AgenticEvolve (No-analysis) & 14 & $\bullet$ & $P (- 0.008724 P + 0.4771 \sin{(0.7761 t )} + 0.8909)$ \\*[3pt]
LLM-SR & 30 & $\circ$ & $P (P (- 2.660 \cdot 10^{-11} \sin{(1.030 t + 0.9963 )} - 0.008724) - 6.197 \cdot 10^{-8} \log{(P )} - 0.4771 \sin{(0.7761 t + 9.425 )} + 0.8909)$ \\*[3pt]
SR-Scientist & 27 & $\circ$ & $- 0.008724 P^{2} + 0.4771 P \sin{(0.7761 t - 0.001458 \sin{(532.4 t - 532.4 )} )} + 0.8909 P$ \\*[3pt]
PySR & 29 & -- & $(P + 3.547) (- 0.3013 \sin{(0.7754 t - 2.253 )} + \sin{(5.756 \sqrt{e^{- t}} )}) - 3.065 \sin{(1.223 \sqrt{P} - 1.431 )}$ \\
\midrule
\multicolumn{4}{@{}l}{\textbf{PO10} (physics), target $\dot v$} \\*
Ground truth & & & $F_0 \sin t - \omega_0^2 (\gamma t + 1) x - \omega_0^2 x^3 - \omega_0^2 x$ \\*[3pt]
AgenticEvolve & 18 & $\bullet$ & $x (- 0.04946 t - 1.992 x^{2} - 3.983) + 0.7998 \sin{(t )}$ \\*[3pt]
AgenticEvolve (No-analysis) & 30 & -- & $- 1.910 x^{3} - 4.929 x + 0.7906 \sin{(1.001 t )} + 0.3278 \cos{(2.400 t )} + 0.2552 \cos{(2.578 t )}$ \\*[3pt]
LLM-SR & 18 & -- & $- 1.849 x^{3} - 4.960 x - 0.7922 \cos{(1.001 t + 1.564 )}$ \\*[3pt]
SR-Scientist & 20 & $\bullet$ & $- 0.04946 t x - 1.992 x^{3} - 3.983 x + 0.7998 \sin{(t )}$ \\*[3pt]
PySR & 24 & -- & $1.984 \log{((\frac{\cos{(x )}}{(t + 24.48) (0.1259 x^{2} + 0.2914)})^{x} )} + 0.8072 \sin{(t )}$ \\
\midrule
\multicolumn{4}{@{}l}{\textbf{PO0} (physics), target $\dot v$} \\*
Ground truth & & & $F_0 \sin t - \beta \sin v - \omega_0^2 x^3 - \omega_0^2 x e^{-|x|}$ \\*[3pt]
AgenticEvolve & 29 & $\circ$ & $- 1.027 x^{3} - 1.027 x e^{- \left|{x}\right|} + 0.9480 \sin{(t )} - 0.7123 \sin{(v )} + 3.792 \cdot 10^{-8} \max(31.12, t) - 1.201 \cdot 10^{-6}$ \\*[3pt]
AgenticEvolve (No-analysis) & 28 & -- & $v (0.1774 \left|{v}\right| - 0.7777) - 0.6113 x \left|{x}\right|^{2.694} - 0.7915 x + 0.9529 \sin{(1.000 t )}$ \\*[3pt]
LLM-SR & 26 & -- & $- 0.5587 v + x (x (- 0.07089 v - 0.8099 x) - 0.5791) - 0.9089 \cos{(0.9994 t + 1.607 )}$ \\*[3pt]
SR-Scientist & 23 & -- & $- 0.6202 v - 0.8070 x^{3} - 0.6018 x + 0.9540 \sin{(t )} + 0.02842 \cos{(t )}$ \\*[3pt]
PySR & 20 & -- & ${0.8228}^{t} \sin{(1.281 t - 0.3232 )} + x (- 0.7675 x^{2} - 0.1472)$ \\
\midrule
\multicolumn{4}{@{}l}{\textbf{PO26} (physics), target $\dot v$} \\*
Ground truth & & & $F_0 \sin t - \beta \log(|v| + 1) - 2 \beta v - \omega_0^2 x^3$ \\*[3pt]
AgenticEvolve & 22 & $\bullet$ & $- 0.8261 v - 2.452 x^{3} - 0.4130 \log{(\left|{v}\right| + 1.000 )} + 0.3403 \sin{(t )}$ \\*[3pt]
AgenticEvolve (No-analysis) & 28 & $\circ$ & $- 0.8261 v + x (3.728 \cdot 10^{-9} - 2.452 x^{2}) - 0.4130 \log{(\left|{v}\right| + 1.000 )} - 0.3403 \cos{(t + 1.571 )}$ \\*[3pt]
LLM-SR & 30 & -- & $v (- 0.4187 v - 0.8830) + x (x (0.2168 - 2.673 x) + 0.05252) + 0.3431 \sin{(1.001 t - 0.1371 )} - 0.07092$ \\*[3pt]
SR-Scientist & 27 & -- & $- 0.8144 v + 0.3140 \sin{(t )} - 6.257 \sin{(x )} + 3.125 \sin{(2 x )} - 0.04366 \cos{(t )} - 0.09358$ \\*[3pt]
PySR & 21 & -- & $(- 0.8261 x + 0.8261 (0.1537 - 0.7009 v) (\sin{(t )} + 0.4212) - 0.1175) e^{{0.6188}^{t}}$ \\
\midrule
\multicolumn{4}{@{}l}{\textbf{PO16} (physics), target $\dot v$} \\*
Ground truth & & & $F_0 \sin t - \beta v \sin x - \beta \sin v - \omega_0^2 x^3$ \\*[3pt]
AgenticEvolve & 29 & $\circ$ & $- 0.1302 v \sin{(x )} - 3.106 x^{3} + 2.423 \cdot 10^{-5} x + 0.8891 \sin{(t )} - 0.1302 \sin{(v )} - 2.430 \cdot 10^{-5} \sin{(x )}$ \\*[3pt]
AgenticEvolve (No-analysis) & 26 & -- & $0.01830 v^{3} + v (- 0.1153 x - 0.1280) - 3.106 x^{3} - 0.8891 \cos{(t - 4.712 )}$ \\*[3pt]
LLM-SR & 22 & -- & $- 0.1032 v - 3.119 x^{3} + 0.01974 x - 0.8918 \cos{(0.9995 t + 7.877 )}$ \\*[3pt]
SR-Scientist & 28 & $\circ$ & $3.152 \cdot 10^{-9} t x - 0.1302 v \sin{(x )} - 3.106 x^{3} + 0.8891 \sin{(t )} - 0.1302 \sin{(v )}$ \\*[3pt]
PySR & 28 & $\bullet$ & $- 0.1302 v \sin{(x )} - 3.106 x^{3} + 0.8891 \sin{(t )} - 0.1302 \sin{(v )}$ \\
\midrule
\multicolumn{4}{@{}l}{\textbf{PO1} (physics), target $\dot v$} \\*
Ground truth & & & $F_0 \sin t - \omega_0^2 x - \omega_0^2 x e^{-|x|}$ \\*[3pt]
AgenticEvolve & 14 & $\bullet$ & $- 1.154 x (1.000 + e^{- \left|{x}\right|}) + 0.4226 \sin{(t )}$ \\*[3pt]
AgenticEvolve (No-analysis) & 29 & -- & $x (1.024 \left|{x}\right|^{0.9231} - 4.398) + 0.4226 \sin{(t )} + 2.335 \sin{(0.8757 x )} + 0.01210 \sin{(2.945 x )}$ \\*[3pt]
LLM-SR & 25 & -- & $0.08775 x^{3} - 1.670 x - 0.1033 \sin{(3.120 x )} + 0.4243 \sin{(1.000 t - 0.002382 )}$ \\*[3pt]
SR-Scientist & 14 & -- & $0.09518 x^{3} - 1.699 x + 0.4630 \sin{(t )}$ \\*[3pt]
PySR & 26 & -- & $- 1.033 x + 0.4226 (1.896 (\sin^{2}{(x )})^{0.3065} - 0.7913) \sin{(x )} + 0.4226 \sin{(t )} - 1.033 \sin{(x )}$ \\
\midrule
\multicolumn{4}{@{}l}{\textbf{PO7} (physics), target $\dot v$} \\*
Ground truth & & & $-\alpha v^3 - \beta |v|^{1/3} - \omega_0^2 x^3$ \\*[3pt]
AgenticEvolve & 18 & $\bullet$ & $- 0.04589 v^{3} - 1.283 x^{3} - 0.7296 \left|{v}\right|^{0.3333}$ \\*[3pt]
AgenticEvolve (No-analysis) & 18 & $\bullet$ & $- 0.04589 v^{3} - 1.283 x^{3} - 0.7296 \left|{v}\right|^{0.3333}$ \\*[3pt]
LLM-SR & 30 & -- & $v (- 0.1387 v - 0.1209) + x (x (0.1172 - 1.242 x) - 0.1226) + 0.2421 \cos{(1.075 t - 5.013 )} - 0.5707$ \\*[3pt]
SR-Scientist & 15 & -- & $- 0.03471 v - 1.466 x^{3} + 0.3496 x - 0.5619$ \\*[3pt]
PySR & 28 & -- & $- 8.558 x + 1.037 (- 0.04107 v^{3} + 0.04107 x - 0.7098) (v^{2})^{0.1713} + 8.558 \sin{(x )}$ \\
\midrule
\multicolumn{4}{@{}l}{\textbf{MatSci11} (material science), target $\sigma$} \\*
Ground truth & & & $a \varepsilon^{p} e^{-Q/(RT)} + b \varepsilon^3 (T - T_0)$ \\*[3pt]
Reduced truth & & & $a \varepsilon^{p} e^{-k/(273 + 500 \varepsilon)} + c_3 \varepsilon^3 + c_4 \varepsilon^4$ \\*[3pt]
AgenticEvolve & 23 & -- & $60.06 \varepsilon^{0.4776} + 2.464 \varepsilon^{1.451} - 538.5 \varepsilon^{2.987} + 3812 \varepsilon^{4.001}$ \\*[3pt]
AgenticEvolve (No-analysis) & 29 & -- & $(1204 \sqrt{\varepsilon} + \varepsilon (\varepsilon (2.636 \cdot 10^{4} \varepsilon - 1.250 \cdot 10^{4}) + 2736) - 888.7) e^{- \frac{1914}{T + 397.4}} + \frac{1.399 \cdot 10^{4}}{T}$ \\*[3pt]
LLM-SR & 27 & -- & $T (T (0.0001784 T - 0.07866) + 5.313) - 425.9 \varepsilon + \varepsilon^{1.622} (8880 - 59.00 T) + 784.0$ \\*[3pt]
SR-Scientist & 23 & -- & $60.06 \varepsilon^{0.4776} + 2.464 \varepsilon^{1.451} - 538.5 \varepsilon^{2.987} + 3812 \varepsilon^{4.001}$ \\*[3pt]
PySR & 27 & -- & $(4.118 e^{7.776 \varepsilon} - 1.635 \cos{(19.57 \varepsilon )} - \cos{(9.543 e^{\varepsilon} )} + 9.987)^{\cos{(e^{- 25.68 \varepsilon} )}}$ \\
\midrule
\multicolumn{4}{@{}l}{\textbf{MatSci9} (material science), target $\sigma$} \\*
Ground truth & & & $c_1 \varepsilon^3 - c_2 (T - T_0) + c_3 \varepsilon^3 (T - T_0)$ \\*[3pt]
Reduced truth & & & $c_0 + c_1 \varepsilon + c_3 \varepsilon^3 + c_4 \varepsilon^4$ \\*[3pt]
AgenticEvolve & 25 & $\circ$ & $T \varepsilon^{2} (8.638 \varepsilon - 1.951) + \varepsilon (\varepsilon (532.7 - 1688 \varepsilon) - 195.1) + 14.00$ \\*[3pt]
AgenticEvolve (No-analysis) & 15 & $\bullet$ & $\varepsilon (\varepsilon^{2} (4319 \varepsilon - 305.5) - 195.1) + 14.00$ \\*[3pt]
LLM-SR & 17 & $\circ$ & $(8.638 T - 2664) (1.930 \cdot 10^{-10} T + \varepsilon^{3.000} - 0.04518) + 0.1922$ \\*[3pt]
SR-Scientist & 17 & $\bullet$ & $4319 \varepsilon^{4} - 305.5 \varepsilon^{3} - 195.1 \varepsilon + 14.00$ \\*[3pt]
PySR & 26 & -- & $- 0.4041 T + 1857 \sin^{3.728}{(\varepsilon )} + 5716 \sin^{5.311}{(\varepsilon )} - 0.07926 \cos{(24.98 \varepsilon )} + 124.6$ \\
\midrule
\multicolumn{4}{@{}l}{\textbf{MatSci6} (material science), target $\sigma$} \\*
Ground truth & & & $c_1 \varepsilon \left(1 - c_2 (T - T_0)\right) + c_3 \varepsilon (T - T_0)^2$ \\*[3pt]
Reduced truth & & & $c_1 \varepsilon + c_2 \varepsilon^2 + c_3 \varepsilon^3$ \\*[3pt]
AgenticEvolve & 27 & $\circ$ & $\varepsilon (\varepsilon (10.43 \varepsilon - 19.84) + 5.857) + (T - 273.0) (\varepsilon (2825 \varepsilon - 787.3) + 54.90) + 3.221 \cdot 10^{-5}$ \\*[3pt]
AgenticEvolve (No-analysis) & 17 & $\bullet$ & $\varepsilon (412.7 T + 1.412 \cdot 10^{6} \varepsilon^{2} - 6.000 \cdot 10^{5} \varepsilon - 8.521 \cdot 10^{4})$ \\*[3pt]
LLM-SR & 19 & $\circ$ & $(3.255 T - 888.5) (0.06180 T + 867.9 \varepsilon^{2} - 272.8 \varepsilon)$ \\*[3pt]
SR-Scientist & 15 & $\bullet$ & $1.412 \cdot 10^{6} \varepsilon^{3} - 3.937 \cdot 10^{5} \varepsilon^{2} + 2.746 \cdot 10^{4} \varepsilon$ \\*[3pt]
PySR & 23 & -- & $- 1.909 T (\sqrt{\varepsilon} + \cos{(1.636 \sqrt{T} )}) + (T \sin{(3.519 \varepsilon^{2} )})^{1.909}$ \\
\end{longtable}

}

\subsection{Effect of the Final Selection}
In seven runs the candidate set contains an exact equation, but the selected equation has negligible extra terms. Five are AgenticEvolve runs on CRK35, PO0, PO16, MatSci9 and MatSci6. In each of these runs, the exact equation is the smallest candidate evaluated and lies on the Pareto front. The PySR score measures the decrease in log error per added node relative to the next smaller candidate on the front. The smallest candidate has no smaller member for comparison and is assigned a score of zero. All candidates on these fronts have training errors at the level of numerical round-off and within 1.5 times the minimum error, so all are eligible for selection. Larger candidates with slightly lower errors receive positive scores, and the candidate with the highest score is selected. On CRK35, for example, the exact equation has 10 nodes and an NMSE of $2.126\times10^{-15}$, while the selected equation has 16 nodes and an NMSE of $2.123\times10^{-15}$.

In the other two runs, SR-Scientist on BPG18 and LLM-SR on MatSci6, the smallest exact equation is excluded because its training error exceeds 1.5 times the minimum error in that run. The LLM-SR run also contains a larger exact equation that meets the error criterion but has a lower score than the selected equation.

In these seven runs the selection rule affects only whether the selected equation is exact or has negligible extra terms.

\subsection{Accuracy and Resources}
Table~\ref{tab:sr-metrics} gives the in-domain and out-of-domain error of every selected equation in the symbolic-regression comparison. Table~\ref{tab:sr-budget} gives the token composition, cost and wall time of every method.

\begin{table*}[t]
\centering
\caption{Numerical accuracy of the selected equation per problem: normalized mean squared error on the in-domain (ID) and out-of-domain (OOD) test sets, computed locally with the shared scorer. Bold: accuracy within tolerance $10^{-3}$ on that set.}
\label{tab:sr-metrics}
\footnotesize
\begin{tabular}{lrrrrrrrrrr}
\toprule
Problem & \multicolumn{2}{c}{AgenticEvolve} & \multicolumn{2}{c}{\begin{tabular}[b]{@{}c@{}}AgenticEvolve\\(No-analysis)\end{tabular}} & \multicolumn{2}{c}{LLM-SR} & \multicolumn{2}{c}{SR-Scientist} & \multicolumn{2}{c}{PySR} \\
 & ID & OOD & ID & OOD & ID & OOD & ID & OOD & ID & OOD \\
\midrule
CRK21 & \textbf{1.4e-15} & \textbf{6.8e-15} & 5.9e-09 & 2.0e-05 & 3.1e-09 & 1.2e-05 & 9.8e-10 & \textbf{5.6e-07} & 1.1e-07 & 4.8e-05 \\
CRK35 & \textbf{2.0e-15} & \textbf{3.0e-14} & \textbf{2.0e-15} & \textbf{2.9e-14} & 2.6e-08 & 9.2e-05 & \textbf{2.0e-15} & \textbf{2.9e-14} & 9.6e-06 & 3.8e-03 \\
BPG18 & \textbf{1.2e-12} & \textbf{1.7e-12} & \textbf{1.2e-12} & \textbf{1.7e-12} & \textbf{1.2e-12} & \textbf{1.7e-12} & \textbf{2.1e-14} & \textbf{2.2e-14} & 2.8e-03 & 3.4e-03 \\
PO10 & \textbf{7.4e-15} & \textbf{1.4e-14} & 1.8e-03 & 1.8e-02 & 6.0e-03 & 2.0e-02 & \textbf{7.4e-15} & \textbf{1.4e-14} & 9.4e-05 & 5.5e-04 \\
PO0 & \textbf{1.8e-13} & \textbf{6.8e-13} & 8.9e-05 & 7.2e-05 & 8.8e-04 & 6.9e-04 & 1.5e-03 & 1.2e-03 & 1.0e-02 & 1.1e-02 \\
PO26 & \textbf{1.5e-13} & \textbf{1.0e-12} & \textbf{1.5e-13} & \textbf{1.0e-12} & 1.9e-03 & 1.5e-03 & 6.6e-03 & 1.1e-02 & 2.9e-02 & 3.2e-02 \\
PO16 & \textbf{4.9e-14} & \textbf{1.3e-13} & 7.4e-06 & 5.8e-06 & 1.6e-03 & 1.7e-03 & \textbf{4.9e-14} & \textbf{1.3e-13} & \textbf{3.5e-10} & \textbf{3.7e-10} \\
PO1 & \textbf{1.3e-14} & \textbf{9.6e-14} & \textbf{9.7e-10} & \textbf{2.3e-09} & 8.5e-05 & 2.6e-04 & 2.3e-03 & 5.4e-03 & 1.8e-06 & 6.9e-07 \\
PO7 & \textbf{8.2e-15} & \textbf{1.1e-14} & \textbf{8.2e-15} & \textbf{1.1e-14} & 4.3e-03 & 8.2e-03 & 1.2e-02 & 3.2e-02 & 3.9e-04 & 5.8e-04 \\
MatSci11 & \textbf{1.4e-14} & \textbf{1.4e-12} & \textbf{1.4e-10} & \textbf{4.0e-07} & 8.9e-07 & 3.3e-04 & \textbf{1.4e-14} & \textbf{1.4e-12} & 2.2e-05 & 7.1e-02 \\
MatSci9 & \textbf{2.7e-14} & \textbf{8.1e-13} & \textbf{3.2e-14} & \textbf{1.0e-12} & \textbf{2.9e-14} & \textbf{8.1e-13} & \textbf{3.2e-14} & \textbf{1.0e-12} & 6.8e-07 & 6.2e-04 \\
MatSci6 & \textbf{1.6e-14} & \textbf{9.0e-13} & \textbf{1.5e-14} & \textbf{8.9e-13} & \textbf{1.7e-14} & \textbf{9.3e-13} & \textbf{1.6e-14} & \textbf{1.1e-12} & 3.4e-05 & 1.6e-01 \\
\midrule
Acc$_{0.001}$ passes & 12 & 12 & 8 & 8 & 3 & 3 & 7 & 8 & 1 & 1 \\
\bottomrule
\end{tabular}
\end{table*}

\begin{table*}[t]
\centering
\caption{Model usage of the symbolic-regression runs, totals over the twelve problems: completed requests, input tokens served from the provider's cache, uncached input tokens, output tokens (of which reasoning), estimated cost at the DeepSeek-V4.1-Flash prices of the main text, and end-to-end wall time. Wall time depends on machine load and API latency and is reported for reference only.}
\label{tab:sr-budget}
\footnotesize
\begin{tabular}{lrrrrrrr}
\toprule
Method & Requests & Cached input (M) & Uncached input (M) & Output (M) & Reasoning (M) & Cost (\$) & Wall time (min) \\
\midrule
AgenticEvolve & 4939 & 568.9 & 2.4 & 7.4 & 6.0 & 6.54 & 313 \\
AgenticEvolve (No-analysis) & 7809 & 1232.5 & 5.6 & 12.6 & 11.0 & 12.12 & 432 \\
LLM-SR & 12000 & 3.2 & 3.7 & 184.6 & 183.3 & 111.34 & 4521 \\
SR-Scientist & 5496 & 131.6 & 2.9 & 9.6 & 7.3 & 6.59 & 808 \\
PySR & -- & -- & -- & -- & -- & -- & 8 \\
\bottomrule
\end{tabular}
\end{table*}

\section{Evaluation-Access Audits}\label{supp:sec:audits}
Candidate programs and analysis scripts run as ordinary host processes without an operating-system sandbox (Section~\ref{sec:discussion}). Access restrictions are therefore part of the protocol, and compliance was checked after the runs. Each audit examined the recorded code, evaluation feedback, analysis output and model reasoning without executing any recorded programs. Pattern searches flagged file access, introspection, references to host paths and output of computed values. Every flagged item was reviewed manually together with the model's next turn.

\subsection{Heuristic Design Runs}
All heuristic runs predate the feedback restriction of Section~\ref{sec:protocol}. Failed evaluations returned truncated output and the full traceback to the model. Successful evaluations returned only metrics. OpenEvolve received the same error text, which appeared in a prompt only when the failed program was later sampled as a parent. The audit covers all 90 reported runs and 13 earlier attempts that were repeated after infrastructure failures.

Of the 131 failed evaluations in the reported runs, 120 exposed the evaluator's absolute path in the traceback. No candidate program or analysis script referred to this path, read or listed a file outside its temporary directory, or used introspection or subprocesses. No model reasoning trace mentioned the evaluator file or the hidden-instance files.

Two No-analysis runs on OBP deliberately used failed evaluations to return computed values. In the DeepSeek run, island 0 first tested whether an exception message would be returned. It then simulated about 150 rules on self-generated sequences at import time and returned their rankings in exception messages. In the GLM run, the program returned its internal counters in an exception message at the last call of the formal evaluation. Neither returned item sizes or other hidden-instance data. Printed output was never returned, although 156 No-analysis programs on OBP contain print statements.

Three AgenticEvolve analyses attempted to reconstruct the hidden instances from the returned metrics by scanning random seeds or decoding rule R1. All three failed. One GLM run on OBP claimed to reproduce the hidden sequences from a particular seed, but only 1.1\% of its item values matched the hidden sequences, consistent with chance.

No heuristic run used the evaluator path, read files outside its temporary directory or reconstructed the hidden instances.

\subsection{Symbolic Regression Runs}
AgenticEvolve, LLM-SR, SR-Scientist and PySR ran on framework version 0.2.0. The reported No-analysis runs used version 0.2.1, which introduced the feedback restriction and a check when loading candidates. This check rejects candidates that import modules other than NumPy and math, call file or code-execution functions, or use double-underscore names.

The evaluator calls every candidate with symbolic inputs and fits the resulting expression itself, so candidates never receive data values. They can access data only by reading files or process memory. The test sets and ground truth are stored in the benchmark's dataset cache on the host and are absent from all task and run directories. Table~\ref{tab:audit-sr} lists the data each method's code is permitted to use and its feedback on evaluation failure. PySR executes no model-written code and was excluded from the audit.

\begin{table}[h]
    \centering
    \caption{Data available to the model-written code of each symbolic-regression method and feedback of a failed evaluation in the reported runs.}
    \label{tab:audit-sr}
    \small
    \begin{tabular}{lll}
        \toprule
        Method & Data available to model-written code & Feedback of a failed evaluation \\
        \midrule
        AgenticEvolve (v0.2.0) & training data, in analysis scripts & output and full traceback \\
        AgenticEvolve (No-analysis, v0.2.1) & none & one-line summary \\
        LLM-SR & none & none \\
        SR-Scientist & training data, in its analysis tool & last line of the error output \\
        \bottomrule
    \end{tabular}
\end{table}

The audit covers 316 AgenticEvolve evaluations and 1857 analyses, 3988 No-analysis evaluations, 12012 LLM-SR samples, and 66 SR-Scientist candidates with 5211 executions of its analysis tool. A further 210 analysis calls and 5 evaluator calls requested by SR-Scientist were never executed because the round had reached its message limit.

Fourteen commits in six No-analysis runs attempted to obtain values from the evaluation. Thirteen used print statements or exceptions to return markers, input statistics or parameters. One also attempted to load the training file at import time from three guessed relative paths. None returned data: the inputs were symbolic, printed output was withheld, and the evaluation directory contained no data file. After each attempt, the agent noted that no output had been returned and resumed the search. No BPG18 commit read or listed a file. The frequency on BPG18 was recovered through successive trial values guided by the returned NMSE (Section~\ref{sec:sr}).

No AgenticEvolve candidate program accessed a file. In five runs, analysis scripts searched their own directory for test data or metadata about the ground truth and found only the training data. On PO0, a failed evaluation exposed the evaluator's absolute path in a traceback. An analysis script then searched for the evaluator's source code under the host's home and temporary directories and tried to import the evaluator, despite the system prompt's instruction to stay within its directory. Neither attempt succeeded, and the agent did not try again.

No LLM-SR sample imports modules, performs file or process operations, uses introspection or outputs values. Samples execute in a forked child of the driver process, which holds the current problem's test data in memory. No sample accesses the interpreter internals through which those data could be reached. SR-Scientist's analysis code reads the training data from standard input as instructed, and neither its analyses nor its candidates read files or refer to host paths.

The original No-analysis runs under version 0.2.0 were replaced. On BPG18, a candidate on island 2 raised an error whose traceback revealed the evaluator's absolute path. Subsequent candidates on that island searched the task directory, loaded the training file and returned the results of frequency scans and least-squares fits in exception messages. The recovered frequency reached the other two islands through migration. Candidates in six other runs tried relative paths or listed directories without finding the file. These attempts prompted the feedback restriction and candidate load check introduced in version 0.2.1.

In the final runs, no candidate program or analysis script obtained data beyond what its method was given. This conclusion is based on the recorded code, outputs and reasoning. The audit cannot exclude effects absent from these records. For LLM-SR, the records contain only the extracted function bodies, not the full model responses.

\end{document}